\documentclass{article}
\usepackage{graphicx}
\usepackage{amsmath}
\DeclareMathOperator*{\argmin}{arg\,min}
 \usepackage[preprint]{neurips_2026}

\usepackage[utf8]{inputenc} 
\usepackage[T1]{fontenc}    
\usepackage{url}            
\usepackage{booktabs}       
\usepackage{amsfonts}       
\usepackage{nicefrac}       
\usepackage{microtype}      
\usepackage{xcolor}         
\usepackage[hidelinks]{hyperref}

\title{Prediction horizon shapes representations\newline in predictive learning}

\author{%
  Aviv Ratzon \& Omri Barak \\
  Technion – Israel Institute of Technology\\
  \texttt{aviv.ratzon@hotmail.com} \\
}

\begin{document}
\maketitle
\begin{abstract}
  Predictive learning has emerged as a central paradigm for training models across diverse data domains and is increasingly viewed as a foundation for modern artificial intelligence. A common intuition for this success is that accurate prediction requires models to capture the underlying dynamics of the environment, leading to the emergence of structured “world models.” However, predictive learning does not universally yield such representations, and a mechanistic account of when and why it does remains incomplete. In this work, we identify the prediction horizon as a critical, but often implicit, component of predictive learning objectives. We show that increasing the prediction horizon fundamentally shapes the effective structure of the learning problem. In a minimal setting, we demonstrate both theoretically and empirically that the model's implicit biases interact with this structural change to recover the latent geometry of the task. We then extend these empirical results to nonlinear architectures and more complex datasets, where similar phenomena persist. These findings provide a principled explanation for the emergence of structured representations in predictive learning paradigms and clarify the conditions under which such representations should be expected.
\end{abstract}

\section{Introduction}

Predictive learning has emerged as a powerful framework for training models that generalize well across a wide range of domains. In its simplest form, the task is minimally specified: a model is trained to predict the next observation given the current one. Despite this simplicity, predictive objectives have driven major advances across domains including language, vision, video, and audio \citep{brown2020language,chen2020generative,sun2019videobert,baevski2020wav2vec2}. In neuroscience, they have also been used to study how networks acquire \emph{world models}—internal representations that capture the latent structure of the environment—by requiring them to predict future inputs given the past observations \citep{recanatesi2021predictive, stachenfeld2017hippocampus, levenstein2024sequential}. These representations are often hypothesized to underlie the success of predictive learning, as accurate prediction necessitates capturing the underlying dynamics of the data.

Despite these advances, several fundamental questions remain unresolved. While many studies report that predictive learning networks develop low-dimensional, interpretable representations of latent variables, this outcome is not guaranteed. Intuitively, learning a latent world model seems advantageous for predictive tasks, as structured representations may enable more efficient predictions. However, overparameterized networks admit infinitely many solutions that can achieve perfect performance without forming any interpretable representation \citep{frankle2018lottery, zhang2016understanding, nguyen2017loss}. The mere success of predictive learning at minimizing prediction error therefore does not explain why structured world models emerge in practice. A deeper understanding of the inductive biases introduced by prediction horizon, network architecture, and training dynamics is needed to clarify \emph{when} and \emph{why} predictive learning induces meaningful internal representations.

We begin by noting that a predictive learning task cannot be specified without specifying the prediction horizon. In this work, we take a first step toward understanding the effect of varying this horizon, and the broader principles governing predictive representations, by constructing a minimal linear predictive learning problem inspired by previous work  \cite{recanatesi2021predictive}. This setting is analytically tractable, yet rich enough to capture essential aspects of the phenomenon. We show empirically that when the network is sufficiently deep and the prediction horizon scales linearly with environment size, gradient descent consistently converges to highly structured solutions that recover the underlying state from the observations. To explain this behavior, we combine tools from machine learning theory with an analysis of the structure of the ordinary least squares (OLS) estimator, revealing how prediction horizon and loss choice shape the solution landscape.

Building on this intuition, we then extend our study to more complex settings, including nonlinear networks, continuous environments with stochastic observations, and settings with multiple independent environments. Across these experiments, we test the generality of the principles uncovered in the linear case, exploring how task structure, training biases, and prediction horizon interact to produce world models.

Our contributions are threefold:  
\begin{enumerate}
\item \textbf{Empirical characterization of horizon-induced representation learning in a minimal predictive task.} In an analytically tractable linear setting, we show that predictive learning does not generically recover latent state structure, but does so reliably when the prediction horizon is sufficiently large relative to the environment size and the network is sufficiently deep. We systematically characterize how prediction horizon and depth shape the geometry of the learned representations.

\item \textbf{A mechanistic account linking prediction horizon, estimator structure, and implicit bias.} We make a connection that was previously overlooked: in deep linear neural networks trained for multiclass classification, the parameters converge to a hard-margin solution with regularization over the rank structure of the effective weight matrix rather than its $L_2$ norm. We then relate the effect of increasing prediction horizon to changes in the structure and spectrum of the OLS estimator. Together, these results provide a concrete explanation for how prediction horizon biases models toward ordered representations aligned with the latent state in the linear, abstract setting.

\item \textbf{Generalization beyond the linear toy model.} We extend the analysis empirically to nonlinear and more naturalistic settings, including continuous environments, stochastic observations, multiple independent environments, non-Euclidean geometry, and predictive coding networks. Across these settings, we find the same qualitative pattern: longer prediction horizons robustly promote low-dimensional representations that reflect the underlying latent geometry.

\end{enumerate}

Together, these results advance our understanding of how predictive learning interacts with model architecture, task design, and optimization dynamics to shape internal representations.

\section{Related Work}

Predictive learning has been shown to uncover latent structure in environments \cite{recanatesi2021predictive}. In particular, predictive learning can recover low-dimensional latent spaces in discrete, continuous, and angular settings \cite{recanatesi2021predictive}. However, this prior work did not examine when such structure fails to emerge. More recent work extended this line of research by showing that recurrent networks form continuous attractors under large-horizon, but not next-step, prediction, highlighting the importance of prediction horizon \cite{levenstein2024sequential}. Our study builds on these findings by analyzing how horizon length, network depth, and optimization dynamics bias predictive learning solutions.

Previous work has proposed several approaches for extracting latent structure within predictive frameworks. One line of work introduced models that enforce locally linear latent dynamics through explicit architectural priors \cite{watter2015embed}. Other work encouraged simplified latent dynamics by imposing a soft state-invariance regularizer, biasing the latent state to change slowly unless driven by actions \cite{saanum2024simplifying}. Related approaches have also learned structured latent transitions using contrastive objectives that separate true next states from negatives \cite{kipf2019contrastive}. While these methods demonstrate that predictive learning can reveal aspects of latent geometry, they rely on architectural constraints, explicit regularization, or assumed structure in the environment. In contrast, our work provides a mechanistic explanation for why and when predictive learning alone, without explicit regularization, reshapes the data geometry and consistently drives networks toward representations that recover the underlying latent state.

Separately, theoretical results on implicit bias in classification show that gradient descent converges to the hard-margin SVM solution for linearly separable data. This has been established for single-layer \citep{soudry2018implicit}, deep linear \citep{ji2018gradient}, and homogeneous networks \citep{lyu2019gradient}.

Finally, while our findings are related to Neural Collapse \cite{papyan2020prevalence}, they are qualitatively different: in our setting, representations collapse toward the latent geometry of the environment rather than toward a simplex, indicating that the effect is driven by environmental structure rather than by an optimal decoding geometry.

\section{Results}

\begin{figure}[h]
\begin{center}
\includegraphics[width=1\textwidth]{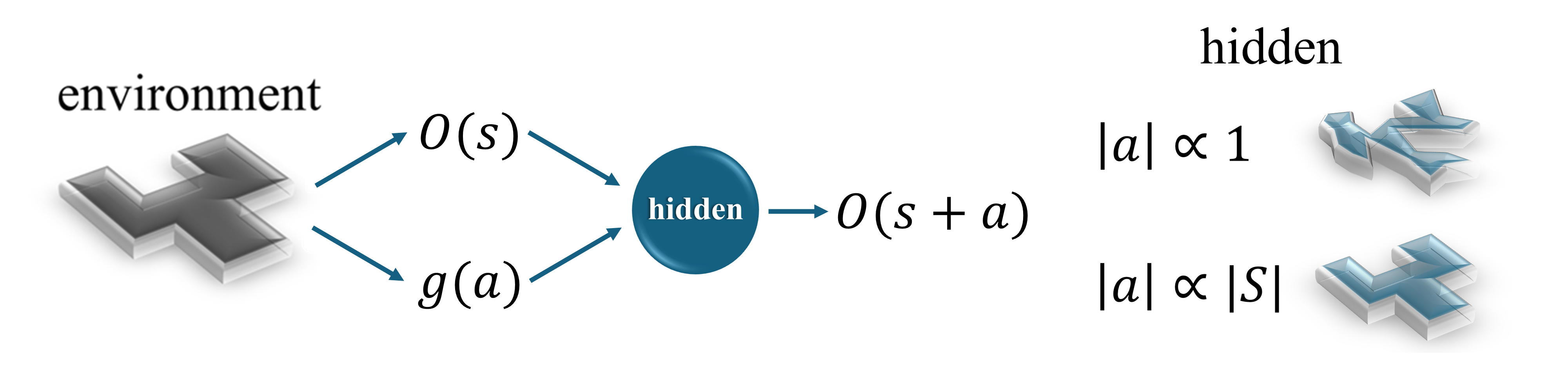}
\end{center}
\caption{Illustration of predictive learning setting where time is abstracted away. An agent is acting in an environment, producing a set of observations and actions in its trajectory. The task is to predict, for each action and observation pair, the following observation. The environment has an underlying structure, and training a model on a predictive learning task sometimes generates a representation of this latent structure. Recent work has shown that increasing the prediction horizon can lead to more accurate and stable representations \citep{levenstein2024sequential}.}
\label{fig1}
\end{figure}

We begin by formalizing the task of predictive learning in a simple setting (Figure~\ref{fig1}).  

\[
\mathbf{x} = (O(s), g(a)), \quad \mathbf{y} = O(s+a), \quad s \in \{1,...,S\}, \; a \in \{-A,...,A\}.
\]

Here, $O$ and $g$ denote high-dimensional observation functions parameterized by lower-dimensional state and action variables, $s$ and $a$. $S$ denotes the number of states, and $A$ the maximal action range, interpreted as the model's prediction horizon. This formulation abstracts away explicit time dependence: rather than modeling continuous temporal evolution, it samples from a memory buffer of state--action trajectories. In the appendix, we show that the same qualitative results persist when actions are provided sequentially rather than encoded by magnitude (Figure~\ref{fig_seq}), indicating that compositional state--action representations can recover state-space structure regardless of input encoding. We begin with a deterministic, discrete environment in which each state maps to a unique observation; these assumptions are relaxed in more complex and naturalistic settings.

\subsection{Spontaneous collapse to order in abstract predictive learning}

We consider an abstract predictive learning task inspired by previous work \cite{recanatesi2021predictive}, in which $S$ states and $2A+1$ actions are represented by one-hot encoded observations $O(s)=\delta_s \in \mathbb{R}^S$, $s \in [1,S]$, and $g(a)=\delta_{a+A} \in \mathbb{R}^{2A+1}$, $a \in [-A,A]$. The network receives $\{O(s), g(a)\}$ and predicts $O(s+a)$, discarding tuples that map to undefined states. We train a bias-free deep linear network with $L$ layers on all valid state--action pairs. Since observations are one-hot encoded, Cross Entropy (CE) loss naturally casts the task as multiclass classification.

Our goal is to study the relationship between the environment's latent geometry and the network's internal representation. The former is given by the shifted state $s+a$, while the latter is defined as the activation of the last hidden layer. Crucially, the use of one-hot encoding removes any intrinsic correlations between neighboring states. Thus, any emergent structure in the hidden representations must arise from learning the predictive task rather than from structure already present in the inputs.

\begin{figure}[h]
\begin{center}
\includegraphics[width=1\textwidth]{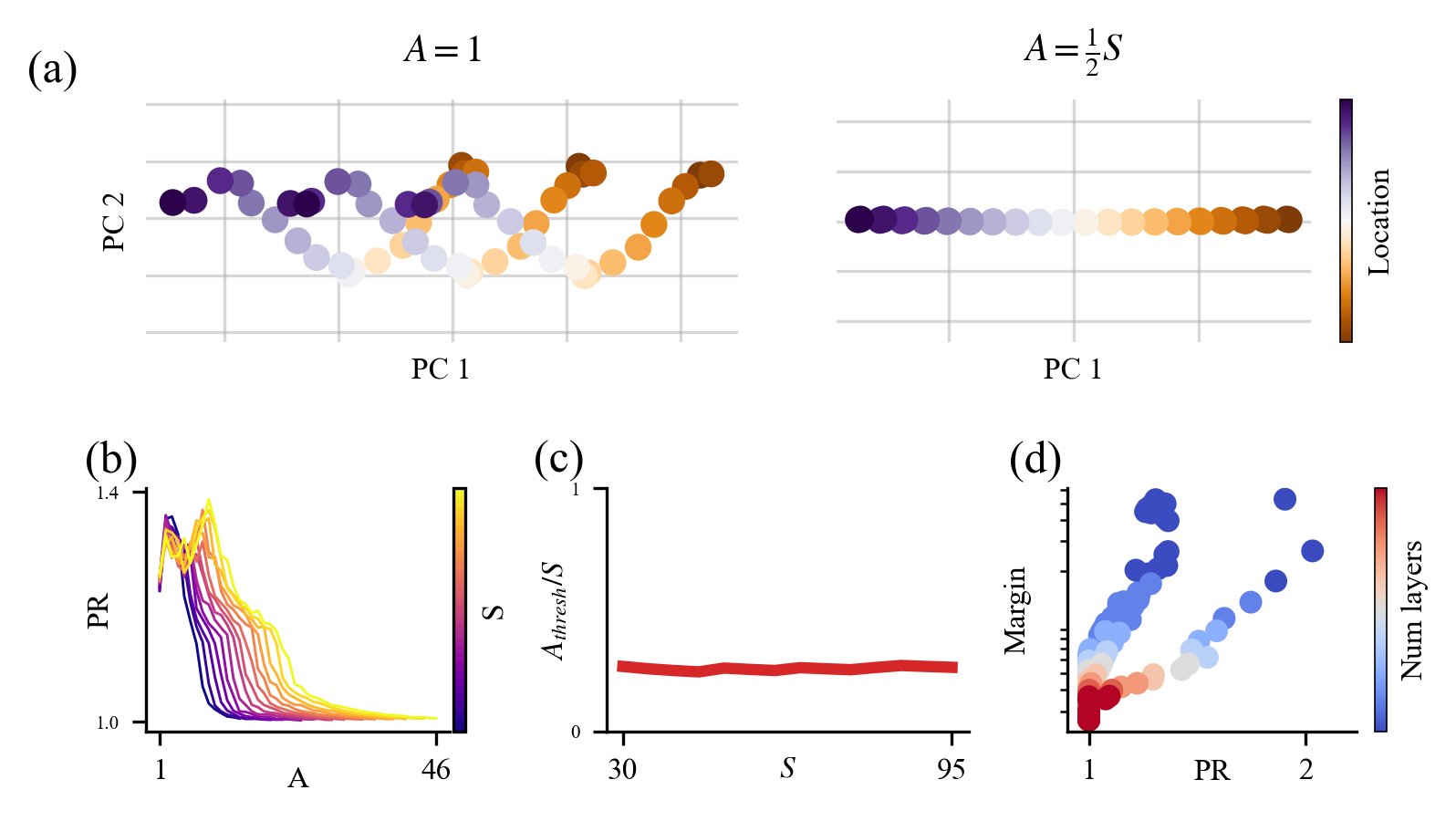}
\end{center}
\caption{(a) The first two principal components of the last hidden layer activations for networks trained on short horizon prediction (left) and large horizon prediction with maximal action $A = S/2$ (right). (b, c): We swept across values of $S$ and $A$  and trained deep linear networks on the predictive task with CE loss. For each $S$ value we plotted the participation ratio (PR) as a function of $A$. We define $A_{\mathrm{thresh}}$ as the smallest horizon for which PR drops 90\% of the total change. As shown, $A_{\mathrm{thresh}}$ scales linearly with $S$, supporting the claim that the prediction horizon required for latent state extraction grows proportionally with the environment size. Figure~\ref{fig2_sup} repeats the same experiment with nonlinear networks. (d) Margins and PR metrics for all networks over all prediction horizons, colored by network depth. } 
\label{fig2}
\end{figure}

In an overparameterized network, there are infinitely many perfect-accuracy solutions for any given $A$. For example, a solution for a large $A$ trivially satisfies the task for smaller $A$, since the latter is a subset of the former. The key question is therefore which solution the network converges to, and why. Figure~\ref{fig2}a shows that for a large prediction horizon, the last hidden-layer activations spontaneously collapse onto the latent state manifold, whereas in the short-horizon setting they do not. The emergence of consistent and highly ordered representations in an unconstrained optimization problem points to strong implicit biases in the training dynamics. Although disorganized solutions are equally valid, the network reliably converges to the collapsed manifold structure---but only when the prediction horizon scales linearly with $S$  (Figure~\ref{fig2} b,c). Note that throughout the paper we use Participation Ratio (PR) to measure dimensionality; see the appendix for details. As we show below, explaining this phenomenon requires combining several theoretical results with a deeper analysis of the problem structure.

\subsection{Mechanism of collapse to order}

The task is multiclass classification with a linear network. Prior work showed that such training converges to a maximal-margin solution \citep{soudry2018implicit, ji2018gradient, lyu2019gradient}. However, as shown in Figure~\ref{fig2}d, networks with varying depths trained on the same task attain different functional margins. This seems to contradict the theory. To resolve this apparent discrepancy, we turn to the concept of \emph{representation cost} \citep{dai2021representation}.

Specifically, \cite{lyu2019gradient} showed that a deep linear network trained on multiclass data converges to parameters that solve
\[
\argmin_{W_1,\dots,W_L} \sum_{l=1}^L |W_l|_2^2
\quad \text{s.t.} \quad
\forall i,\ \forall k \neq y_i:\
W_{y_i}^\top \mathbf{x}_i \geq W_k^\top \mathbf{x}_i + 1,
\]

where $W = \prod_{l=1}^L W_l$. Although this resembles the hard-margin multiclass SVM, it is not identical. The authors of \cite{dai2021representation} further showed that, for deep linear networks, $L_2$ regularization on the parameters induces a Schatten $2/L$ quasi-norm in function space. The corresponding optimization problem in function space is therefore

\[
\begin{aligned}
\argmin_{W} \|W\|^{SC}_{2/L}
&\quad \text{s.t.} \quad 
\forall i,\ \forall k \neq y_i :\ 
W_{y_i}^\top \mathbf{x}_i \geq W_k^\top \mathbf{x}_i + 1, \\
\|W\|^{SC}_{2/L} 
&= \left(\sum_i \sigma_i^{2/L}\right)^{L/2}
\end{aligned}
\]

where $\sigma_i$ are singular values of $W$. This characterization implies that increasing network depth induces a trade-off between effective rank and functional margin: deeper networks are biased toward lower-rank approximations of the hard-margin solution that compromise margins. This explains the empirical observation that both effective rank and functional margin decrease with depth (Figure~\ref{fig2}d)

\begin{figure}[h]
\begin{center}
\includegraphics[width=1\textwidth]{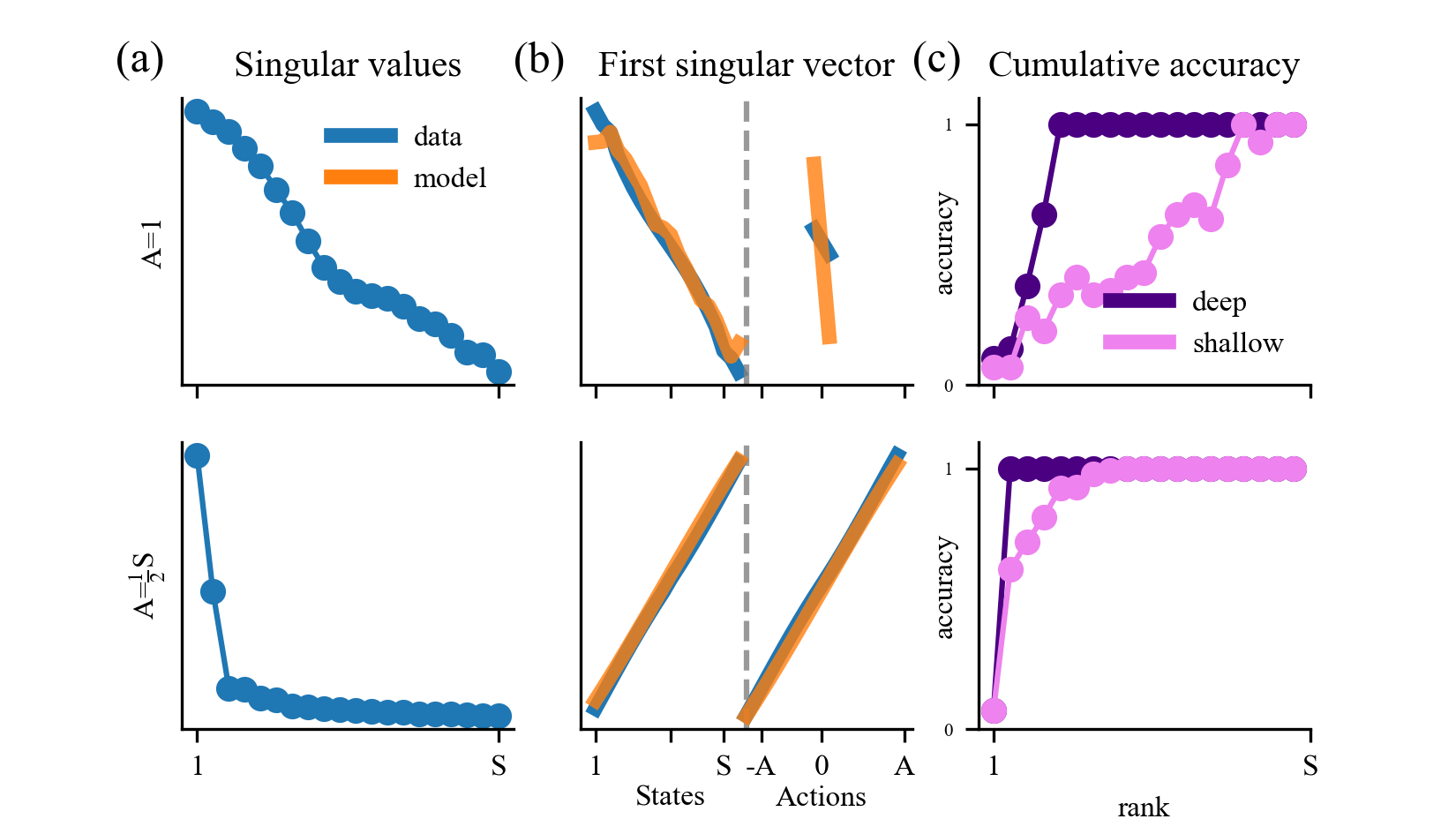}
\end{center}
\caption{Comparison of the spectra of the OLS estimator $\Sigma$ and the effective weight matrix $W$ of trained linear networks under short- and long-horizon predictive tasks. (a) Singular value spectrum of the OLS matrix $\Sigma$. (b) First singular vector of the OLS estimator and the trained model. (c) Comparison between deep (10-layer) and shallow (1-layer) linear networks, showing accuracy as a function of the rank-truncated approximation of $W$. Overall, the figure shows that increasing the prediction horizon makes both the OLS solution and the learned network effectively lower-dimensional, and that deeper networks rely more strongly on the dominant low-rank directions. }
\label{fig3}
\end{figure}

At this point, we understand that deep networks are biased toward low-rank approximations of the maximal-margin solution. This still does not explain, however, the dependence on prediction horizon. Increasing the prediction horizon changes the structure of the data: both the input dimension and the number of samples grow with $A$. To gain insight into this change, we study the OLS estimator

\[
\Sigma = (X^\top X)^{\dagger} X^\top Y,
\]

where $(\cdot)^\dagger$ denotes the Moore–Penrose inverse. This matrix captures the linear relation between the inputs and the targets. In highly balanced and symmetric settings such as ours, it also provides intuition about classification, since the OLS solution can separate classes. As shown in Figure~\ref{fig3}, for larger values of $A$ the OLS matrix becomes effectively lower-dimensional and develops two dominant singular values. Its leading singular vector corresponds to a linear transformation that extracts the scalar input state and action, and adds them together. Importantly, as the prediction horizon increases, this direction explains an increasing fraction of the variance, making it progressively more useful for classification. As a result, networks consistently converge to solutions that exploit this direction.

The origin of this low-dimensionality is the approximate band structure of the OLS matrix. As $A$ increases, this band widens, corresponding to lower Fourier modes. Unfortunately, we were unable to prove this statement in our setting. But we were able to derive an analytic formula for dimensionality  in a simpler setting - with cyclic boundary conditions. Although in the cyclic setting the data are not linearly separable, we saw the same qualitative phenomenon in nonlinear networks trained in this setting (Figure~\ref{fig_circ}). Quantitatively, the dimensionality of the OLS matrix has a similar decline with $A$ for both boundary conditions (Figure~\ref{fig_OLS_comp}). We provide several estimations of this dimensionality, the simplest of which is: 

\[
    PR(A, S) = \frac{(\sum_i \sigma_i)^2}{\sum_i \sigma_i^2} = \frac{S^3}{\left(2A+1\right)\left(S^2+2AS+S-4A^2-4A-1\right)},
\]

where $\sigma_i$ are the singular values. For large $S$ we get:
\[
\begin{aligned}
    PR(A \propto 1, S\gg1) \propto S, \\
    PR(A \propto S, S\gg1) \propto \frac{S}{A}.
\end{aligned}
\]

This accounts for the linear scaling of the horizon required to observe collapse.

Because directions that explain most of the variance are also the most useful for class separation, the dominant singular vector of $\Sigma$ naturally becomes the most informative feature for solving the predictive task. This explains why, in the large-horizon regime, deep networks with low-rank bias consistently converge to this leading direction, yielding ordered and low-dimensional representations of the environment. In the short-horizon regime, by contrast, the variance is spread more evenly across directions, and no single direction is sufficiently informative to dominate the solution.

\begin{figure}[h]
\begin{center}
\includegraphics[width=1\textwidth]{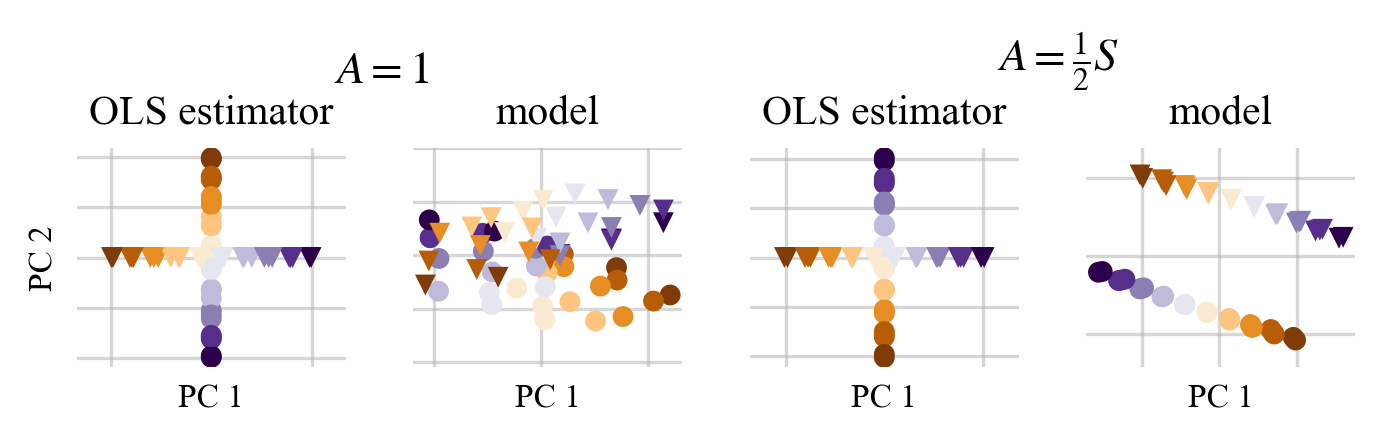}
\end{center}
\caption{Results for the task with two independent environments. Data are projected onto the first two singular vectors of the OLS estimator, confirming that observations and actions from each environment are orthogonal. In the short horizon setting, the model representations remain unaligned, whereas in the large horizon setting they collapse into a shared low-rank structure that aligns the two environments and reveals their underlying symmetry. The shapes (circles, triangles) represent the different environments. }
\label{fig_two_corridors}
\end{figure}

We can now ask whether this mechanism provides intuition beyond the one-dimensional linear setting. Consider, for example, a predictive task with two distinct environments, each with its own encoding of states and actions. In that case, the singular vectors of the OLS estimator associated with the two environments are orthogonal. Because deep networks exhibit a low-rank bias, training should favor solutions that align the two representations in a way that minimizes the number of active singular values. Figure~\ref{fig_two_corridors} illustrates this phenomenon. In both settings, the network embeds the two environments in a shared representation space, but under the large-horizon objective the resulting geometry captures the underlying structure and introduces a symmetry between analogous representation objects. This picture extends beyond one-dimensional latent variables, as illustrated by the two-dimensional example in Figure~\ref{fig_2d}.

One way to interpret the linear mechanism above in a broader setting is through a growing complexity gap between two classes of solutions: structured solutions that recover the latent state, and unstructured solutions that merely memorize the input--output relation. As emphasized in recent theoretical work on world-model learning \cite{zhang2025neural}, these classes can be indistinguishable at the level of training error, since in overparameterized models both can fit the training data perfectly, while differing substantially in functional complexity. From this perspective, the key question is not only whether a solution exists, but which class of solution is preferred by the model's implicit bias. In our setting, increasing the prediction horizon naturally shifts this balance. A larger horizon increases the number of state--action pairs that must be fit, making rote input--output matching progressively more complex. At the same time, these additional samples are not arbitrary: they remain coupled by the transition structure of the prediction problem. The OLS analysis makes this concrete in the linear case, where increasing $A$ reshapes the estimator spectrum and concentrates an increasing fraction of the explainable variance onto a small number of structured directions aligned with the latent geometry. This suggests a broader picture in which longer prediction horizons do not simply provide more supervision, but selectively amplify the simplicity advantage of world-model-like solutions over unstructured memorization. In this sense, our linear results should not be viewed as a universal characterization of predictive learning, but as a concrete instance of a more general principle: increasing prediction horizon can enlarge the set of conditions under which implicit bias drives learning toward latent representations rather than toward direct input--output matching.

\subsection{Generalization across settings}

Next, we consider a more naturalistic setting in which the environment is continuous and observations are temporally correlated. Consider an agent moving between several rooms. Within each room, the environment changes smoothly, whereas transitions between rooms introduce discontinuities. A useful representation of such an environment should use prediction to stitch the rooms together into a coherent world map. In this setup, states are drawn from a uniform distribution, $s \sim U(-1,1)$, and actions are drawn from a Gaussian distribution, $a \sim \mathcal{N}(0,\sqrt{A})$. Figure 5 shows such a world with four rooms, captured by $O(s)$. We train deep nonlinear networks on the task with different prediction horizons. For small $A$, the network merely mirrors the local autocorrelation structure of the data, whereas for larger horizons it stitches together the different linear segments into a coherent one-dimensional manifold. The full technical details are presented in the appendix.

\begin{figure}[h]
\begin{center}
\includegraphics[width=0.75\textwidth]{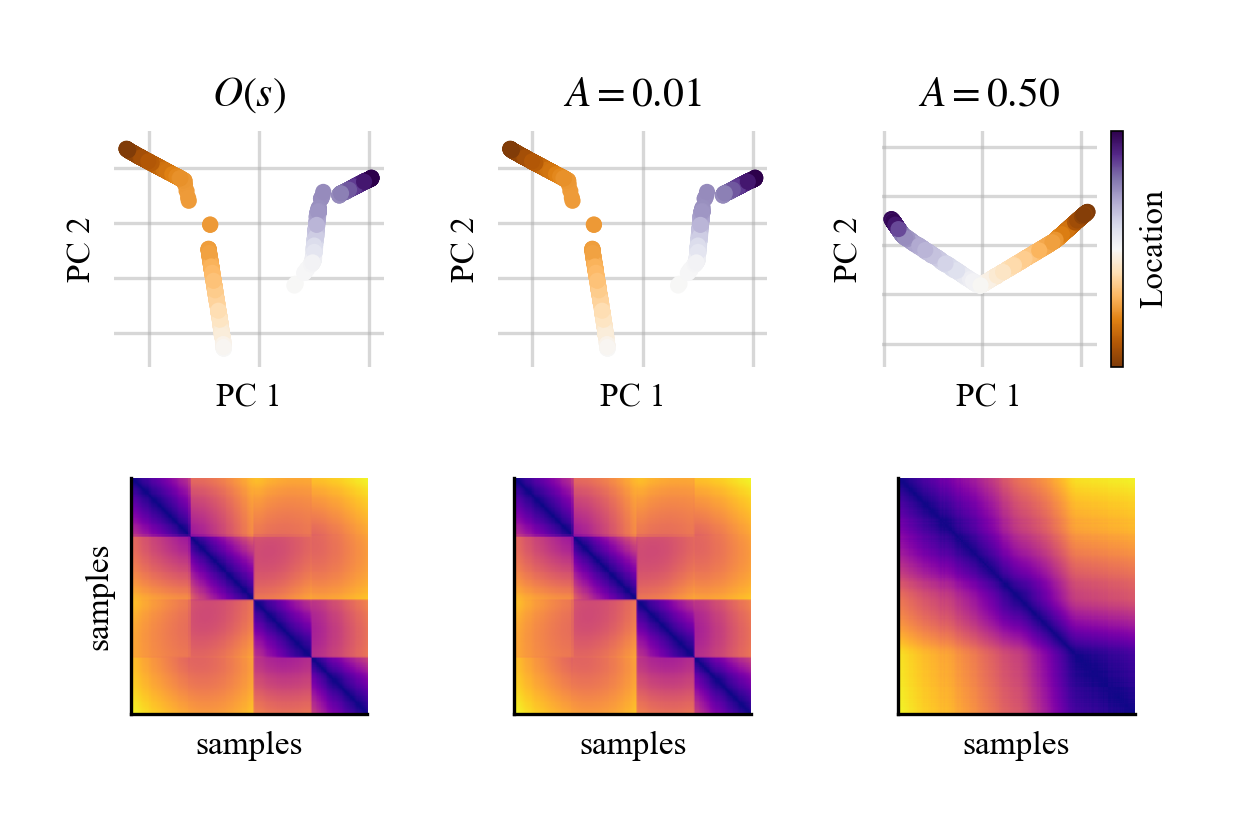}
\end{center}
\caption{Training a deep nonlinear network on a predictive task with observations generated from a piecewise linear function containing three discontinuities. When the action distribution is narrow, the learned representations primarily mirror local autocorrelation. In contrast, with a wider action distribution, the network organizes its hidden representations along a smooth one-dimensional manifold that bridges the discontinuities, thereby recovering the underlying latent state. The top figure shows the Principal Component Analysis (PCA) space, and the bottom figure shows the distance matrix sorted by the state variable.}
\label{fig_piecwise}
\end{figure}

To test whether these observations extend to settings with stochastic rather than deterministic observations, we designed a variant of the MNIST \cite{lecun2002gradient} image generation task. We trained a conditional deep convolutional generative adversarial network (cDCGAN) to generate MNIST images, where the conditioning input consists of an MNIST digit (observation) together with a one-hot encoded digit label (action). The target output is an MNIST image whose class corresponds to the sum of the input digits. As in the abstract setting, the target class is not provided explicitly, and we control the maximal allowed action. To obtain a more objective measure of representation quality, we exclude a subset of state–action pairs during training and evaluate the model on these held-out combinations, yielding out-of-distribution (OOD) test data. Correct predictions on these pairs require the model to learn a compositional representation of states and actions. In addition, we analyze the latent space of the condition encoder to assess the linearity of the learned representations. Full experimental details are provided in the appendix.

\begin{figure}[h]
\begin{center}
\includegraphics[width=1\textwidth]{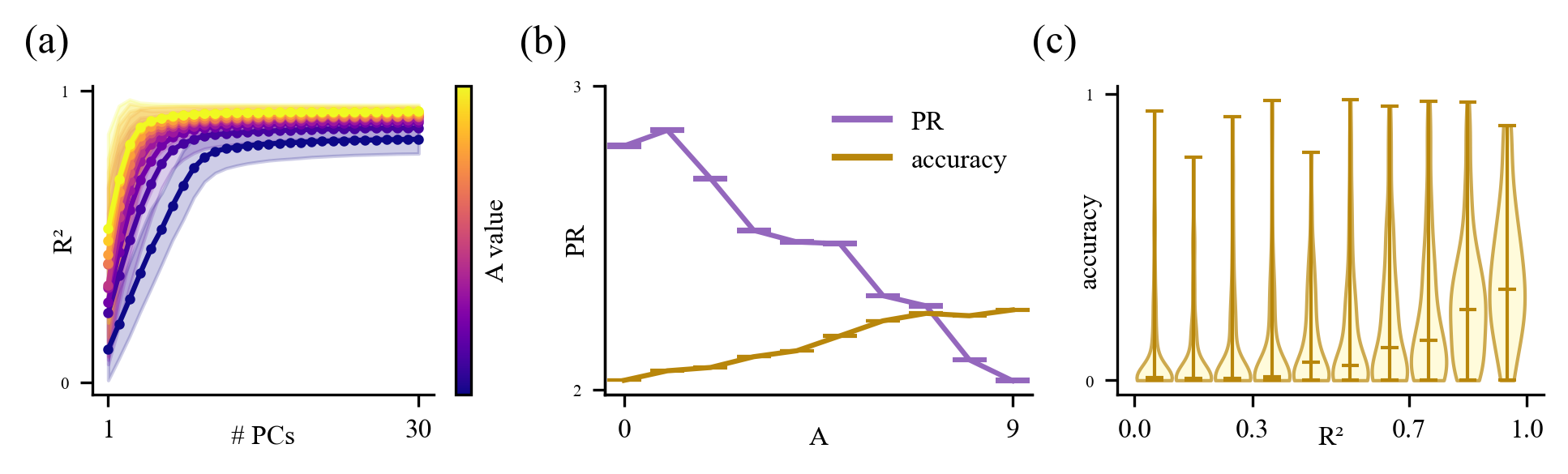}
\end{center}
\caption{Results of the MNIST experiments. (a) Mean $R^2$ between cumulative PCs of the encoder last layer activity and the target digit value. (b) PR of last layer activity and accuracy over OOD test samples as a function of the maximal action $A$ used during training. Accuracy was measured using a pre-trained classifier which was tested over the generated images. Error bars show Standard Error of Mean. We note that distributions are wide and full violin plots are shown in the appendix (Figure~\ref{fig_mnist_sup}). (c) Distribution of accuracy over OOD test samples as a function of the linear decodability of the target digit from the encoder activity. Overall the results show that for larger prediction horizon representations become more linear, and these are correlated with better generalization. }
\label{fig_MNIST}
\end{figure}

We trained multiple cDCGANs across different prediction horizons and analyzed their behavior. In both short- and long-horizon settings, the models generate visually correct digits (see appendix). However, increasing the prediction horizon leads to systematically lower-dimensional representations from which the target state is more linearly decodable (Figure~\ref{fig_MNIST}). Consistent with this, models trained with larger horizons achieve higher accuracy on the excluded state–action pairs, and this OOD performance correlates with the linear decodability of the target label. These results indicate that longer-horizon predictive learning promotes structured and interpretable representations even in this more naturalistic setting.

\begin{figure*}[t]
\centering
\begin{minipage}[t]{0.49\textwidth}
    \centering
    \includegraphics[width=\textwidth]{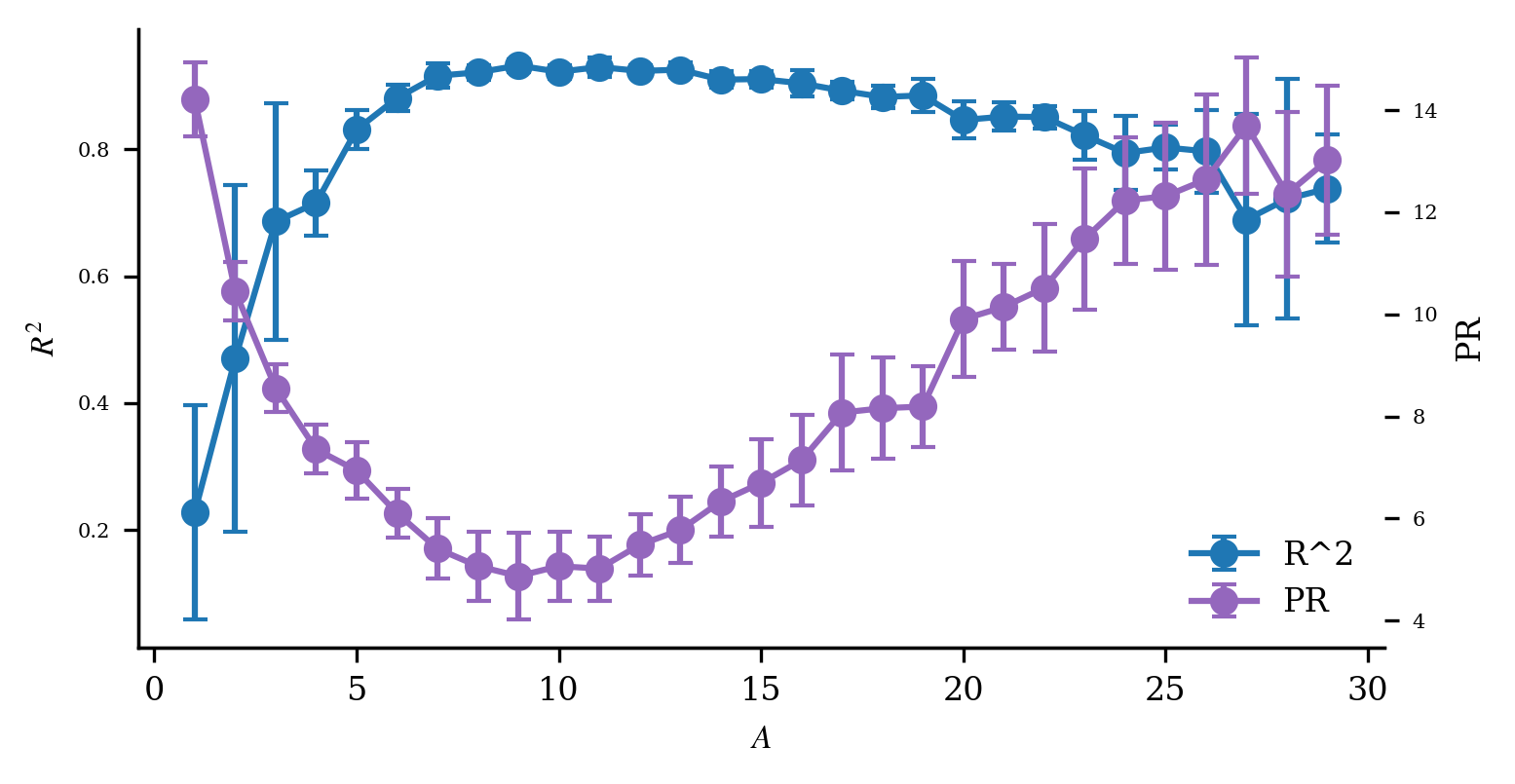}
    \caption{Predictive Coding Network trained on the discrete state abstract prediction task. As prediction horizon increases, the dimensionality reduces and linear decodability of the latent from the first PC increases. Note that there is an ``optimal'' prediction horizon around $A\approx \frac{1}{2}S$, after which dimensionality increases. This phenomenon of an optimal mesoscale prediction horizon has repeated across various settings, but is not explored in this work.}
    \label{fig_pcn}
\end{minipage}
\hfill
\begin{minipage}[t]{0.49\textwidth}
    \centering
    \includegraphics[width=\textwidth]{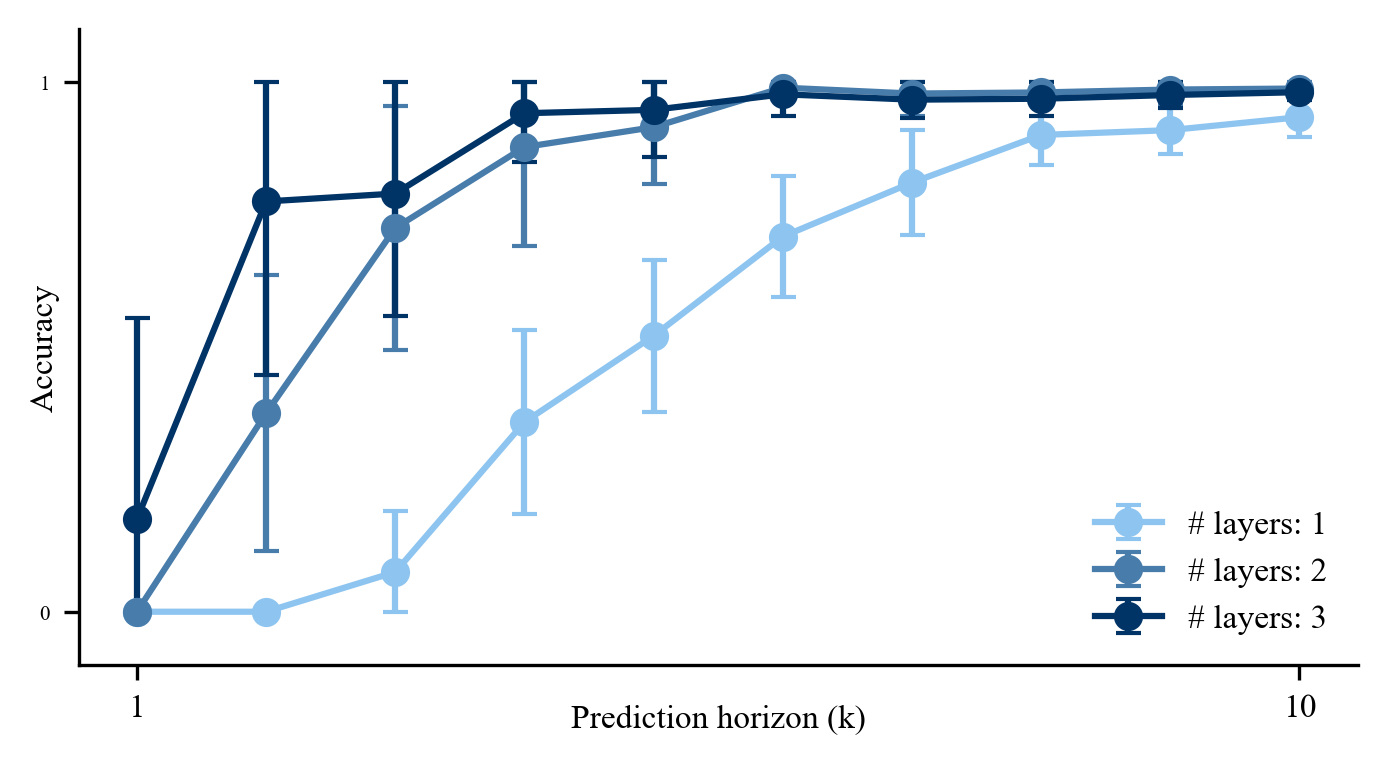}
    \caption{Accuracy as a function of prediction horizon in the tree-geometry setting. Curves correspond to different network depths, and error bars denote the standard deviation across random seed initializations. The results highlight two complementary effects: increasing the prediction horizon reshapes the task by widening the complexity gap between solutions that recover the latent structure and those that rely on unstructured input-output matching, while increasing network depth strengthens the model’s implicit bias toward lower-complexity solutions.}
    \label{fig_tree}
\end{minipage}

\end{figure*}

We now consider a natural setting for this predictive learning task: Predictive Coding Networks (PCNs), which offer a biologically plausible mechanism for such learning. We trained a PCN with 5 layers in the simplified and abstract discrete environment while varying the prediction horizon, observing the same qualitative results as in our previous experiments (Figure~\ref{fig_pcn}). Namely, for short prediction horizons and shallow network depths, representations remain unstructured. Conversely, in deep networks with long prediction horizons, the representations organize along the underlying one-dimensional geometry of the environment. A detailed description of these simulations is provided in the appendix.

Finally, we consider data whose underlying structure is non-Euclidean. In all datasets presented so far, the latent geometry was Euclidean, for which a low-rank implicit bias is well suited. To examine whether our results extend to other geometries and their corresponding inductive biases, we construct a dataset generated from a balanced binary tree. This setting induces a hyperbolic geometry, which is less amenable to direct visualization and standard quantitative metrics. Instead, following the MNIST experiment, we evaluate representation quality using held-out OOD state--action pairs. Successful generalization to these pairs indicates that the model has learned a compositional representation (Figure~\ref{fig_tree}). Full experimental details are provided in the appendix.

\section{Discussion}

We have shown that large-horizon predictive learning, together with the appropriate implicit biases, acts as a strong inductive force that drives networks toward low-dimensional, structured representations of the environment's latent variables. Increasing the prediction horizon makes the task more constrained, revealing a dominant direction in the data to which deep networks, biased toward low-rank solutions, naturally align. This explains why structured solutions consistently emerge in the large-horizon setting even when many trivial solutions are available.

In neuroscience, predictive learning has long been proposed as a model for hippocampal function. Recent experimental \cite{vollan2025left} and theoretical \cite{levenstein2024sequential} work has emphasized the importance of sequential prediction, which, in light of our results, can be understood through the lens of prediction horizon. Viewing experimental findings through abstract tasks such as predictive learning provides an alternative to rigid, mechanistic interpretations of brain function. For instance, rather than treating place cells as components of a hard-coded navigational system, they can be interpreted as emerging from the extraction of statistical regularities in sensory input. From this perspective, the notion of prediction horizon offers a concrete framework for both interpreting existing results and designing new experiments. In particular, the dependence of optimal prediction horizons on environment size suggests experiments involving environments with multiple spatial scales, which could help infer the effective prediction horizon used by the brain.

Several open questions remain. It is still unclear why the dominant direction takes such a strongly structured form and how this intuition extends to more complex nonlinear settings. Our experiments on continuous environments, MNIST, and predictive coding networks suggest that similar principles apply more broadly, but a full theoretical account remains for future work. More generally, these findings suggest that longer prediction horizons may play an important role in the emergence of interpretable world models in both machine learning and neuroscience.

\newpage

\bibliography{references}
\bibliographystyle{unsrt}

\newpage

\appendix

\renewcommand{\thefigure}{S\arabic{figure}}
\setcounter{figure}{0} 

\begin{figure}[h]
\begin{center}
\includegraphics[width=1\textwidth]{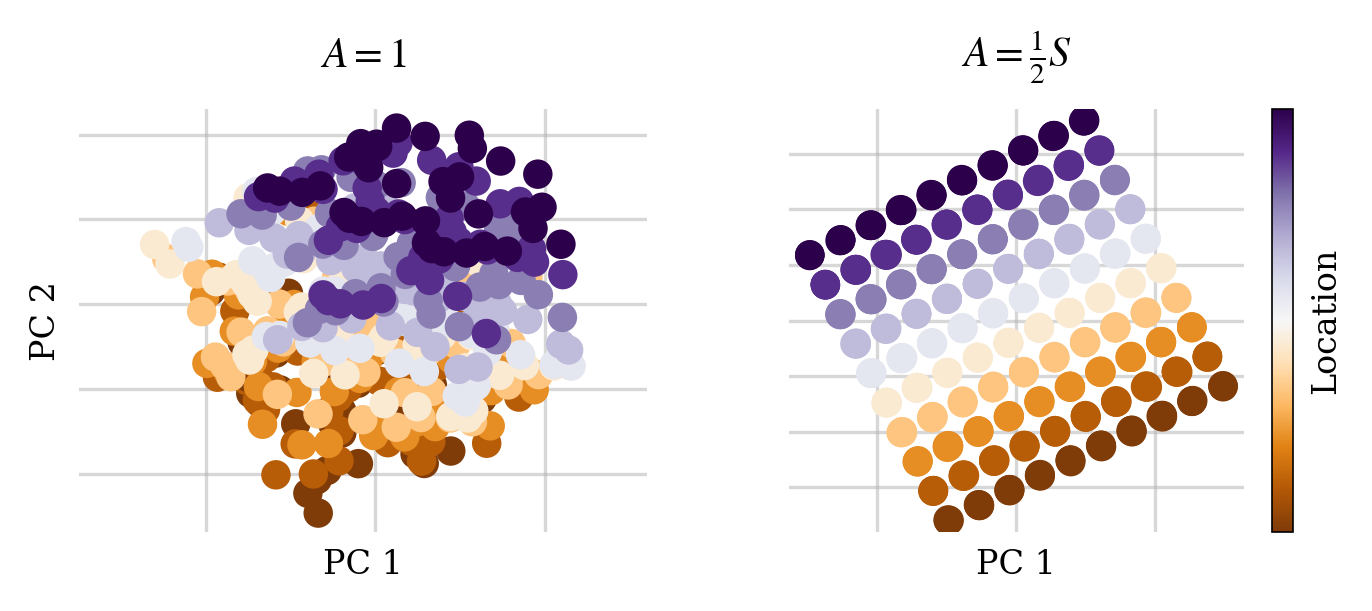}
\end{center}
\caption{Same abstract task as the linear case, but for a two-dimensional state variable.}
\label{fig_2d}
\end{figure}

\begin{figure}[h]
\begin{center}
\includegraphics[width=1\textwidth]{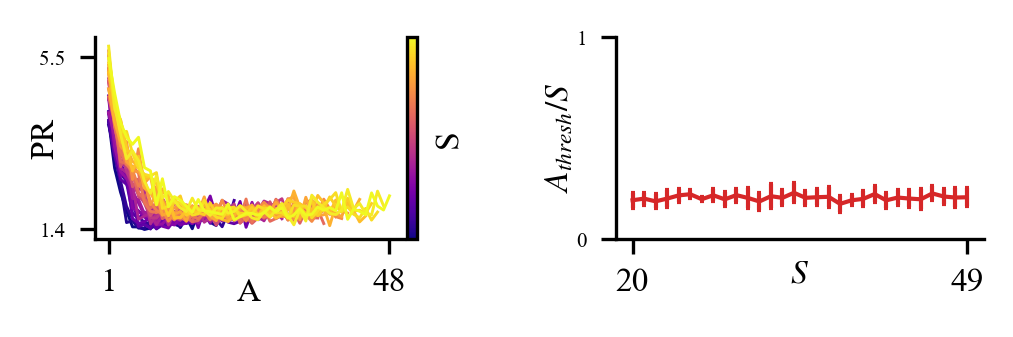}
\end{center}
\caption{Same experiments as Figure~\ref{fig2}, but with ReLU activations and Mean Squared Error loss. Error bars are over random seed initializations.}
\label{fig2_sup}
\end{figure}

\section{Code Availability}
A publicly accessible GitHub repository will be made available upon acceptance. Detailed and sufficient technical specifications required to generate the code for simulations are presented in the appendix.

\section{LLM usage}
LLMs were used to polish the text in the paper, generate code for running simulations, as well as mathematical and technical descriptions in the appendix.

\section{Metrics}

\subsection{Participation Ratio}

We use the participation ratio (PR) to quantify effective linear dimensionality. In the paper, it is used both for hidden activations and for weight matrices.

\paragraph{Hidden activations.}
Let $H \in \mathbb{R}^{M \times N}$ be the hidden activations, with samples $h_i \in \mathbb{R}^N$, and let
\[
\Sigma_H = \frac{1}{M}\sum_{i=1}^M (h_i-\bar{h})(h_i-\bar{h})^\top
\]
be the sample covariance matrix, where $\bar{h}=\frac{1}{M}\sum_{i=1}^M h_i$. If $\lambda_1,\dots,\lambda_N$ are the eigenvalues of $\Sigma_H$, we define
\[
\mathrm{PR}(H)=\frac{\left(\sum_{i=1}^N \lambda_i\right)^2}{\sum_{i=1}^N \lambda_i^2}
=\frac{\mathrm{Tr}(\Sigma_H)^2}{\mathrm{Tr}(\Sigma_H^2)}.
\]
Lower values indicate that the representation is concentrated in fewer directions.

\paragraph{Weight matrices.}
For a matrix $W \in \mathbb{R}^{d_{\mathrm{out}}\times d_{\mathrm{in}}}$ with singular values $\sigma_1,\dots,\sigma_r$, we define
\[
\mathrm{PR}(W)=\frac{\left(\sum_{i=1}^r \sigma_i^2\right)^2}{\sum_{i=1}^r \sigma_i^4}.
\]
This measures the effective rank of $W$: smaller values indicate that only a few singular directions dominate.

\subsection{NC1}
This metric was introduced in \cite{papyan2020prevalence}. Given hidden activations 
\[
h \in \mathbb{R}^{M \times N}, \quad y \in \{1, \dots, C\}^n,
\]
where $M$ is the number of samples, $N$ is the hidden dimension, and $C$ is the number of classes, we define:

\[
\mu = \frac{1}{M} \sum_{i=1}^M h_i
\]
as the global mean of activations, and
\[
\mu_c = \frac{1}{m_c} \sum_{i: y_i = c} h_i, \quad m_c = |\{i : y_i = c\}|
\]
as the class-conditional means.

The within-class scatter matrix is
\[
S_W = \frac{1}{M} \sum_{c=1}^C \sum_{i: y_i=c} (h_i - \mu_c)(h_i - \mu_c)^\top,
\]
and the between-class scatter matrix is
\[
S_B = \frac{1}{M} \sum_{c=1}^C m_c \, (\mu_c - \mu)(\mu_c - \mu)^\top.
\]

The NC1 metric is then defined as
\[
\mathrm{NC1} = \frac{\operatorname{Tr}(S_W)}{\operatorname{Tr}(S_B)}.
\]

\noindent
Intuitively, $S_W$ captures the variance of samples around their respective class means, while $S_B$ captures the variance of class means around the global mean. The ratio $\mathrm{NC1}$ therefore measures the relative tightness of clusters to their separation: smaller values indicate more compact class representations.

\subsection{Margins}
Let $f : \mathbb{R}^d \to \mathbb{R}^C$ denote the network output function, where 
\[
f(x) = (f_1(x), f_2(x), \dots, f_C(x))
\]
are the class scores (logits) for input $x \in \mathbb{R}^d$.  
For each sample $(x_i, y_i)$ with true label $y_i \in \{1, \dots, C\}$, the functional margin is defined as
\[
\gamma_i = f_{y_i}(x_i) - \max_{j \neq y_i} f_j(x_i).
\]

The multiclass margin for the dataset is then given by
\[
\gamma = \min_{i=1,\dots,M} \gamma_i.
\]

\noindent
Intuitively, $\gamma_i$ measures the difference between the score assigned to the correct class and the highest score among all incorrect classes for sample $i$. The overall margin $\gamma$ is the worst-case (smallest) of these values across all samples, and therefore characterizes the minimal separation achieved by the classifier.

\subsection{Order}

In the main text we use a metric defined as the $R^2$ between the first PC and the latent variable. Here we define it in full detail. Let $H \in \mathbb{R}^{M \times N}$ denote the hidden activations, where $M$ is the number of samples and $N$ the hidden dimension.  
We compute the first principal component $\mathbf{u}_1 \in \mathbb{R}^N$ of $H$, i.e. the unit-norm eigenvector of the sample covariance matrix
\[
\Sigma = \frac{1}{M} \sum_{i=1}^M (\mathbf{h}_i - \bar{\mathbf{h}})(\mathbf{h}_i - \bar{\mathbf{h}})^\top
\]
corresponding to the largest eigenvalue, where $\bar{\mathbf{h}} = \frac{1}{M} \sum_{i=1}^M \mathbf{h}_i$.

Each sample is then projected onto this direction:
\[
z_i = \mathbf{u}_1^\top (\mathbf{h}_i - \bar{\mathbf{h}}), \quad i = 1, \dots, M.
\]

Let $s_i \in \mathbb{R}$ denote the state variable associated with sample $i$.  
The \emph{first PC order} metric is defined as the coefficient of determination ($R^2$) of the linear regression between $\{z_i\}$ and $\{s_i\}$:
\[
\text{order} = R^2(z, s).
\]

\noindent
This metric measures how strongly the first principal component of the hidden representations aligns with the state. High values of order indicate that the dominant axis of variation in the representation space reflects the state. For D-dimensional states variables we simply take the first D principal components. 

\subsection{Alignment}
To quantify whether the network aligns representations of separate state variables, we compute an alignment score between subspaces spanned by the leading principal components of their activations. This method is adapted from \cite{sorscher2022neural}.

Let $H^{(1)}, M^{(2)} \in \mathbb{R}^{M \times N}$ denote hidden activations corresponding to two distinct dataset partitions (e.g., two environments or contexts).  
For each partition, we compute the sample covariance
\[
\Sigma^{(k)} = \frac{1}{M} \sum_{i=1}^M \bigl(\mathbf{h}^{(k)}_i - \bar{\mathbf{h}}^{(k)}\bigr)\bigl(\mathbf{h}^{(k)}_i - \bar{\mathbf{h}}^{(k)}\bigr)^\top, 
\quad k \in \{1,2\},
\]
where $\bar{\mathbf{h}}^{(k)}$ is the mean activation of partition $k$.

From $\Sigma^{(k)}$, we extract the top principal directions $U^{(k)} \in \mathbb{R}^{N \times m}$, where $m$ is the smallest number of eigenvectors explaining at least a fixed proportion of variance (e.g., $95\%$).

The alignment score is based on the principal angles $\theta_1 \leq \dots \leq \theta_m$ between the two subspaces spanned by $U^{(1)}$ and $U^{(2)}$. We take the cosine of the smallest principal angle:
\[
\mathrm{Align}(H^{(1)}, H^{(2)}) = \cos(\theta_1).
\]

\noindent
This score lies in $[0,1]$, with higher values indicating stronger alignment (i.e., the leading directions of variability in the two partitions are closely matched). 

\section{Analytical derivation of OLS and PR}
\label{sec:ols_derivation}

\begin{figure}[h]
\begin{center}
\includegraphics[width=0.666\textwidth]{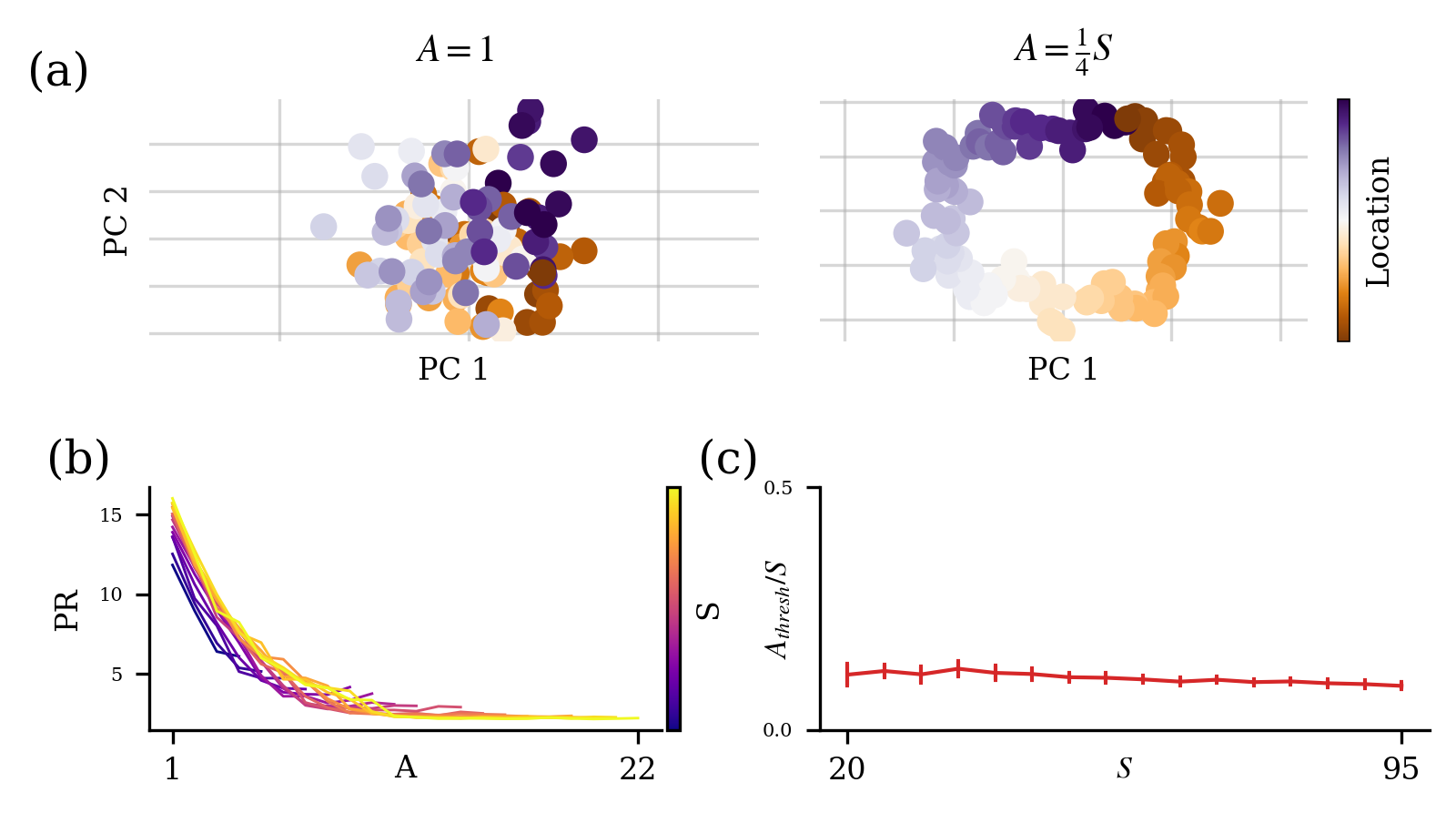}
\end{center}
\caption{Same as Figure~\ref{fig2}, but for deep nonlinear networks and cyclic boundary conditions. Networks were trained with 2 hidden layers, 100 hidden units and CE loss. Values of $A$ were sweeped in the range $1<A<\frac{1}{4}S$ and $20<S<100$. For each parameter combination, simulations were run with 10 random seed initializations.}
\label{fig_circ}
\end{figure}

We seek the Ordinary Least Squares (OLS) estimator $W_{OLS}$ that predicts a one-hot encoded next state from the concatenated one-hot encodings of the current state and action. As noted in the main text, we were only able to obtain an analytical expression under cyclic boundary conditions. We first justify this choice empirically by training deep nonlinear networks on the cyclic task, showing the same qualitative phenomenon: only as the prediction horizon increases do latent representations reflect the underlying geometry and decrease in dimensionality. This dimensionality trend closely follows that of the OLS estimator (Figure~\ref{fig_circ}). The OLS dimensionality is also similar across both boundary conditions in the range $1<A<\frac{1}{4}S$, which is the regime we focus on in our analytical analysis. After $A=\frac{1}{4}S$ states start looping back, and other phenomenon begin to emerge. Additionally, we restrict our analysis to the state block $W_s$, corresponding to the weights on the state-encoding entries. In the cyclic linear setting, the action block is constant, since actions and states cannot be combined linearly. That being said, the main effect is preserved: for large prediction horizons, the latent variable is extracted by the first two singular vectors, which correspond to $\sin$ and $\cos$ of the input state.  

\begin{figure}[h]
\begin{center}
\includegraphics[width=0.666\textwidth]{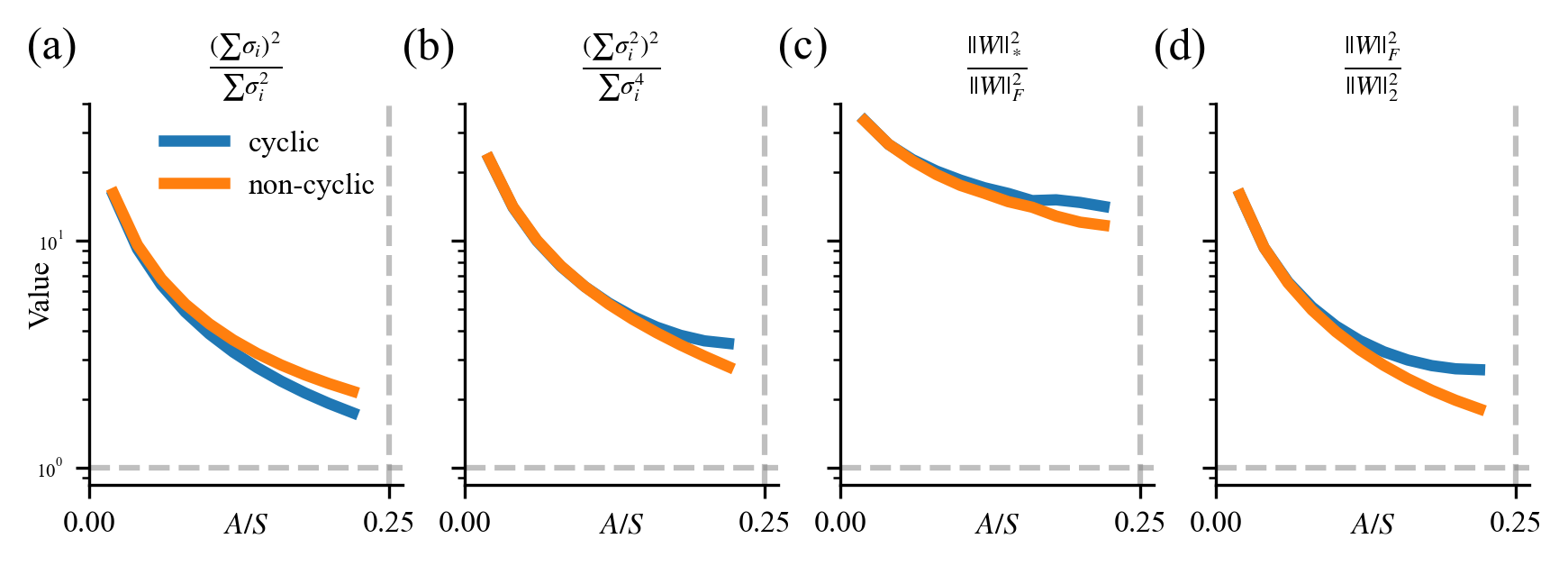}
\end{center}
\caption{Comparing 4 different measures of matrix dimensionality for both cyclic and non-cyclic boundary conditions. (a) PR assuming matrix is PSD. (b) PR assuming matrix is not PSD. (c) PR for non-PSD matrix by taking absolute values of singular values. (d) Soft rank.}
\label{fig_OLS_comp}
\end{figure}

Note that there are several ways to measure the dimensionality of a weight matrix, and Figure~\ref{fig_OLS_comp} compares four such measures empirically, demonstrating that in all cases the OLS exhibits decreased dimensionality in the range $1<A<\frac{1}{4}S$. The main text reports $\frac{(\sum \sigma_i)^2}{\sum \sigma_i^2}$ because it yields the cleanest and most easily interpretable expression. In the following sections, we derive the OLS estimator for the cyclic setting and two expressions for the PR.

\subsection{Derivation of the Closed-Form OLS Estimator}

Let $S$ be the total number of states, and let the set of valid actions be parameterized by bounds $[-A, A]$, giving $m = 2A+1$ total valid actions. The design matrix $X \in \mathbb{R}^{Sm \times (S+m)}$ contains all possible combinations of state and action one-hot vectors. The target matrix $Y \in \mathbb{R}^{Sm \times S}$ contains the corresponding deterministic next states, wrapped cyclically. For simplicity, we restrict our calculation to $A \in \{1,...,\frac{1}{2}S-1\}$.

We wish to find the minimum-norm solution to the normal equations:
\begin{equation}
    (X^T X) W_{OLS} = X^T Y
\end{equation}
Because the input vectors are concatenated one-hot encodings, $X$ is rank-deficient. The unique minimum-norm solution is given by the Moore-Penrose pseudoinverse, $W_{OLS} = (X^T X)^\dagger X^T Y$. We partition the weight matrix into state weights and action weights: $W_{OLS} = [W_S^T, W_A^T]^T$, where $W_S \in \mathbb{R}^{S \times S}$ and $W_A \in \mathbb{R}^{m \times S}$.

Let $e_s \in \mathbb{R}^S$ and $e_a \in \mathbb{R}^m$ be standard basis vectors. Each row of $X$ is $[e_s^T, e_a^T]$. Summing the outer products over all state-action combinations yields the Gram matrix:
\begin{align*}
    X^T X &= \sum_{s=1}^S \sum_{a=1}^m \begin{bmatrix} e_s \\ e_a \end{bmatrix} [e_s^T, e_a^T] \nonumber \\
    &= \sum_{s=1}^S \sum_{a=1}^m \begin{bmatrix} e_s e_s^T & e_s e_a^T \\ e_a e_s^T & e_a e_a^T \end{bmatrix}
\end{align*}
Evaluating these blocks over the sums gives:
\begin{equation*}
    X^T X = \begin{bmatrix} m I_S & J_{S, m} \\ J_{m, S} & S I_m \end{bmatrix}
\end{equation*}
where $I_k$ is the $k \times k$ identity matrix, and $J_{p,q}$ is a $p \times q$ matrix of all ones.

The target matrix $Y$ consists of one-hot encoded next states, $e_{(s+a) \pmod S}^T$. Thus:
\begin{align*}
    X^T Y &= \sum_{s=1}^S \sum_{a=1}^m \begin{bmatrix} e_s \\ e_a \end{bmatrix} e_{(s+a) \pmod S}^T \nonumber \\
    &= \begin{bmatrix} \sum_s e_s \left(\sum_a e_{(s+a) \pmod S}^T\right) \\ \sum_a e_a \left(\sum_s e_{(s+a) \pmod S}^T\right) \end{bmatrix}
\end{align*}
The top block constructs a circulant transition matrix $C \in \mathbb{R}^{S \times S}$, where $C_{s,k} = 1$ if state $k$ is reachable from $s$, and $0$ otherwise. In the bottom block, for any fixed action, summing over all source states hits every target state exactly once due to circularity, yielding a vector of all ones. Therefore:
\begin{equation*}
    X^T Y = \begin{bmatrix} C \\ J_{m, S} \end{bmatrix}
\end{equation*}

We now Solve the Normal Equations under the Null-Space Constraint. Substituting our matrices into the normal equations yields a system of two equations:
\begin{align*}
    m W_S + J_{S, m} W_A &= C  \\
    J_{m, S} W_S + S W_A &= J_{m, S} 
\end{align*}
Let $r_S = \mathbf{1}_S^T W_S$ and $r_A = \mathbf{1}_m^T W_A$ be row vectors representing the column sums of $W_S$ and $W_A$. Using the property $J_{p,q} M = \mathbf{1}_p (\mathbf{1}_q^T M)$, we can rewrite the system as:
\begin{align}
    W_S &= \frac{1}{m}(C - \mathbf{1}_S r_A) \label{eq:ws_iso} \\
    W_A &= \frac{1}{S}(\mathbf{1}_m \mathbf{1}_S^T - \mathbf{1}_m r_S) \label{eq:wa_iso}
\end{align}

Because every input vector $x$ satisfies $\sum_{i=1}^S x_i = 1$ and $\sum_{j=1}^m x_{S+j} = 1$, the design matrix $X$ has a left null space spanned by $v = [\mathbf{1}_S^T, -\mathbf{1}_m^T]^T$. The minimum-norm solution requires that the columns of $W_{OLS}$ are orthogonal to this null space ($v^T W_{OLS} = 0$), which enforces the constraint:
\begin{equation*}
    \mathbf{1}_S^T W_S - \mathbf{1}_m^T W_A = 0 \implies r_S = r_A \equiv r
\end{equation*}

To find $r$, we multiply Equation \ref{eq:ws_iso} by $\mathbf{1}_S^T$. Noting that the column sums of the circular matrix $C$ are all $m$ (hence $\mathbf{1}_S^T C = m \mathbf{1}_S^T$), and $\mathbf{1}_S^T \mathbf{1}_S = S$, we obtain:
\begin{align*}
    r &= \frac{1}{m} (m \mathbf{1}_S^T - S r) \nonumber \\
    r \left(1 + \frac{S}{m}\right) &= \mathbf{1}_S^T \nonumber \\
    r &= \frac{m}{S+m} \mathbf{1}_S^T \label{eq:r_val}
\end{align*}

We can now derive the final closed-form expressions. Substituting $r$ back into Equation \ref{eq:wa_iso} yields the action weights:
\begin{align*}
    W_A &= \frac{1}{S} \mathbf{1}_m \left(\mathbf{1}_S^T - \frac{m}{S+m} \mathbf{1}_S^T\right) \nonumber \\
    &= \frac{1}{S} \mathbf{1}_m \left(\frac{S}{S+m} \mathbf{1}_S^T\right) \nonumber \\
    &= \frac{1}{S+m} J_{m, S}
\end{align*}

Substituting $r$ into Equation \ref{eq:ws_iso} yields the state weights:
\begin{align*}
    W_S &= \frac{1}{m} \left(C - \mathbf{1}_S \left(\frac{m}{S+m} \mathbf{1}_S^T\right)\right) \nonumber \\
    &= \frac{1}{m} C - \frac{1}{S+m} J_{S, S}
\end{align*}

Thus, the complete closed-form minimum-norm OLS estimator is:
\begin{equation*}
    W_{OLS} = \begin{bmatrix} \frac{1}{m} C - \frac{1}{S+m} J_{S, S} \\[1em] \frac{1}{S+m} J_{m, S} \end{bmatrix}
\end{equation*}

\subsection{Approximate participation ratio of the state weights}

We derive the Participation Ratio of the symmetric matrix $W_S$. The Participation Ratio characterizes the effective dimensionality of the eigenspace and is defined as:

\begin{equation*}
    PR = \frac{(\sum \sigma_i)^2}{\sum \sigma_i^2} = \frac{(\text{Tr}(W_S))^2}{\text{Tr}(W_S^2)}
\end{equation*}

Note that this is only correct for a Positive Semi-Definite (PSD) matrix, but since $W_s$ is symmetric and most eigenvalues are positive, we provide this calculation for simplicity. The next section provides a calculation of the precise expression. The state weight matrix is given by $W_S = \frac{1}{m} C - \frac{1}{S+m} J_{S,S}$.

First, we compute the trace of $W_S$. Since the action $a=0$ is always valid, the diagonal elements of the circulant matrix $C$ are all $1$. Thus, $\text{Tr}(C) = S$. The trace of $J_{S,S}$ is also $S$. By linearity:
\begin{equation*}
    \text{Tr}(W_S) = S (\frac{1}{m} - \frac{1}{S+m})
\end{equation*}

Second, because $W_S$ is circulant and symmetric:
\begin{equation*}
    \text{Tr}(W_S^2) = S||w||^2_2 = S(m(\frac{1}{m} - \frac{1}{S+m})^2 + (S-m) \frac{1}{m^2})
\end{equation*}
where $w$ is any row of $W_S$. Substituting both into the PR formula and simplifying we get:
\begin{equation*}
    PR(A, S) = \frac{S^3}{m(S^2+mS-m^2)}
\end{equation*}

And substituting $m=2A+1$:
\begin{equation*}
    PR(A, S) = \frac{S^3}{\left(2A+1\right)\left(S^2+2AS+S-4A^2-4A-1\right)}
\end{equation*}

We can now see how the PR scales for short horizon ($A\propto1$) and larg horizon ($A\propto S$):
\begin{equation*}
\begin{aligned}
    PR(A=1, S\gg1) &= \frac{1}{3}S \\
    PR(A=\alpha S, S\gg1) &= \frac{1}{\alpha+\alpha^2-\alpha^3}
\end{aligned}
\end{equation*}

Demonstration that for short prediction horizon, the OLS has a dimensionality of the order of the environment size, and for large prediction horizon it decreases close to 1 as $A\rightarrow \frac{1}{4}S$.

\subsubsection{Precise participation ratio of the state weights}
\label{sec:pr_w_ols}

We now provide the precise calculation for the PR of the state weights given that the matrix is not PSD:

\begin{equation*}
    PR = \frac{(\sum \sigma_i^2)^2}{\sum \sigma_i^4} = \frac{(\text{Tr}(W_S^2))^2}{\text{Tr}(W_S^4)}
\end{equation*}

The eigenvalues of \(C\) are
\[
c_k
=
\sum_{r=-A}^{A} e^{-2\pi i kr/S},
\qquad k=0,\ldots,S-1.
\]
In particular,
\[
c_0=m.
\]

Since \(J\) has eigenvalue \(S\) on the constant vector and eigenvalue \(0\) on all other Fourier modes, the eigenvalues of \(W\) are
\[
\lambda_0
=
\frac{1}{m}c_0-\frac{S}{S+m}
=
1-\frac{S}{S+m}
=
\frac{m}{S+m},
\]
and for \(k=1,\ldots,S-1\),
\[
\lambda_k
=
\frac{c_k}{m}.
\]

Hence
\[
\operatorname{Tr}(W^4)
=
\lambda_0^4
+
\sum_{k=1}^{S-1} \lambda_k^4
=
\frac{m^4}{(S+m)^4}
+
\frac{1}{m^4}\sum_{k=1}^{S-1} c_k^4.
\]

Using
\[
\sum_{k=0}^{S-1} c_k^4
=
S\cdot \frac{2m^3+m}{3},
\]
we get
\[
\sum_{k=1}^{S-1} c_k^4
=
S\cdot \frac{2m^3+m}{3}
-
m^4.
\]

Therefore,
\[
\operatorname{Tr}(W^4)
=
\frac{m^4}{(S+m)^4}
+
\frac{1}{m^4}
\left(
S\cdot \frac{2m^3+m}{3}
-
m^4
\right).
\]

Simplifying,
\[
\operatorname{Tr}(W^4)
=
\frac{S(2m^3+m)}{3m^4}
-
1
+
\frac{m^4}{(S+m)^4}.
\]

Equivalently,
\[
\operatorname{Tr}(W^4)
=
\frac{S(2m^2+1)}{3m^3}
-
1
+
\frac{m^4}{(S+m)^4}.
\]

Thus,
\[
\boxed{
\operatorname{Tr}(W^4)
=
\frac{S(2m^2+1)}{3m^3}
-
1
+
\frac{m^4}{(S+m)^4}
}
\]

And plugging in both expressions into the PR equation we get:
\[
PR(A, S) = \frac{(S(m(\frac{1}{m} - \frac{1}{S+m})^2 + (S-m) \frac{1}{m^2}))^2}{
\frac{S(2m^2+1)}{3m^3}
-
1
+
\frac{m^4}{(S+m)^4}
}
\]

We can simplify the expression to:

\[
PR(A, S) = \frac{3Sm\left(S^2+Sm-m^2\right)^2}{-10m^6-10Sm^5+m^4+5S^3m^3+4Sm^3+2S^4m^2+6S^2m^2+4S^3m+S^4}
\]

We can now once again see how the PR scales for short horizon ($A\propto1$) and large horizon ($A\propto S$):
\begin{equation*}
\begin{aligned}
    PR(A=1, S\gg1) &= \frac{2}{11}S \\
    PR(A=\alpha S, S\gg1) &= -\frac{3\left(4\alpha^2-2\alpha-1\right)^2}
{4\alpha\left(80\alpha^4+40\alpha^3-5\alpha-1\right)}
\end{aligned}
\end{equation*}

Showing that once again we obtain the same qualitative scaling.

\section{Data Construction for the Piecewise-linear Function Experiment}
\label{app:pwlinear_data_construction}

This section documents the exact dataset-generation procedure used for Figure~\ref{fig_piecwise}.

\paragraph{Sampling of latent state and action.}
For each sample \(i=1,\dots,1000\):
\[
s_i \sim \mathcal U(-1,1), \qquad
a_i \sim \mathcal N(0,\sigma^2),\ \ \sigma=A.
\]
Action $a$ is provided as a scalar feature (not one-hot).
Samples are drawn for each state such that:
\[
-1 \le s_i + a_i \le 1.
\]
This means that the number of samples for each condition is identical.

\paragraph{Piecewise-linear function family.}
Each output coordinate \(j\in\{1,\dots,20\}\) is piecewise linear in \(s\):
\[
f_j(s)=m_{b(s),j}\,s+b_{b(s),j}.
\]
Here:
\begin{itemize}
\item \(m_{k,j}\) and \(b_{k,j}\) are sampled once per run from \(\mathcal U(-2,2)\), there are \(4\) linear segments per dimension. The parameter $k$ denotes segment index.
\item segment assignment uses breakpoints from
\[
\mathrm{linspace}(-1.01,\ 1.01,\ 5).
\]
\end{itemize}
The same sampled piecewise function is used for both \(f(s)\) and \(f(s+a)\) within a run.

\paragraph{Input/target construction.}
Define \(f(s)\in\mathbb R^{20}\). For each valid pair \((s_i,a_i)\):
\[
x_i = [\,f(s_i),\ a_i\,] \in \mathbb R^{21}, \qquad
y_i = f(s_i+a_i)\in\mathbb R^{20}.
\]
Thus the model learns a one-step latent shift under scalar action.

\paragraph{Model architecture.} Since this task is not solvable with a linear network, we use a deep nonlinear network. The network has 5 hidden layers, ReLU activations and 50 units in each layer. Additionally, layers are initialized with xavier normal initialization and a gain of 0.4.

\section{MNIST experiment details}
\paragraph{Task construction.}
We formulate predictive generation over MNIST digits as a conditional transition task. Let the source class be \(s \in \{0,\dots,9\}\), the action be \(a \in [-A, A] \cap \mathbb{Z}\), and the target class be \(t=s+a\). In the non-cyclic setting used in our main runs, samples are restricted to valid transitions with \(0 \le t \le 9\). For each training example, we draw a source image \(x_s\) from class \(s\), sample an action \(a\), and draw an independent target image \(x_t\) from class \(t\). The model is conditioned on \((x_s, \mathrm{onehot}(a))\) and trained to generate an image matching class \(t\). Input images are normalized to \([-1,1]\).

\paragraph{Action encoding and horizon.}
Actions are represented as one-hot vectors over a fixed action space of size \(2(N_{\text{digits}}-1)+1=19\), centered at zero. The horizon parameter \(A\) controls the sampled action range \([-A,A]\). For OOD-style evaluation, we exclude selected state--action pairs from training and evaluate on those held-out combinations. The number of excluded pairs is random, and sampled from  $\mathcal U\{1,A\}$, so simulations with larger prediction horizon also have more held-out pairs. For each value of $A$ we run 300 training procedures with different random seed initialization, resulting in a total of 3000 networks.

\paragraph{Condition encoder.}
The condition encoder maps source image and action one-hot vector to a conditioning embedding \(c \in \mathbb{R}^{128}\).
The image branch is a CNN:
\[
1\times 28\times 28
\rightarrow 32\times 14\times 14
\rightarrow 64\times 7\times 7
\rightarrow 128\times 4\times 4
\rightarrow 256\times 2\times 2,
\]
followed by flattening. The action branch is a two-layer MLP (\(19 \rightarrow 256 \rightarrow 256\)). The concatenated image/action features are processed by a deep MLP (\(512 \rightarrow 256 \rightarrow 256 \rightarrow 256 \rightarrow 256 \rightarrow 128\), with LeakyReLU activations).

\paragraph{Generator and discriminator.}
The generator receives Gaussian noise \(z \sim \mathcal{N}(0,I)\), \(z \in \mathbb{R}^{100}\), concatenated with \(c\). A fully connected stack maps \((z,c)\) to \(256\times 7\times 7\), followed by transposed convolutions:
\[
256\times 7\times 7 \rightarrow 128\times 14\times 14 \rightarrow 64\times 28\times 28 \rightarrow 1\times 28\times 28,
\]
with \(\tanh\) output.
The discriminator is conditional: \(c\) is projected to \(16\) spatial channels and tiled across the image plane, concatenated to the image channels, passed through convolutional blocks, and combined again with \(c\) before a final linear real/fake logit. Spectral normalization is applied to discriminator convolutions and linear layers.

\paragraph{Optimization protocol.}
Training uses BCE-with-logits adversarial loss for 30 epochs, batch size 128, and AdamW (\(\text{lr}=2\times10^{-4}\), \(\beta_1=0.5\), \(\beta_2=0.999\)). The discriminator and generator are updated alternately each batch. The condition encoder parameters are included in \emph{both} optimizers (generator-side and discriminator-side updates). To reduce mode collapse and discriminator overconfidence, we use: (i) one-sided label smoothing for real labels (\(0.9\) vs. fake \(0.0\)), (ii) instance noise added to real and fake discriminator inputs with initial standard deviation \(0.1\), decayed per epoch by factor \(0.99\), and (iii) spectral normalization in the discriminator.

\paragraph{Classifier-based evaluation and representation metrics.}
To quantify semantic correctness, generated images are evaluated with an auxiliary MNIST CNN classifier trained for 10 epochs. We report classifier accuracy on generated samples and, when excluded pairs are defined, held-out-pair accuracy and mean absolute digit error \(|\hat{t}-t|\). To quantify representation geometry of the condition encoder, we compute:
\begin{enumerate}
\item \textbf{Participation Ratio} from covariance eigenvalues of encoder last layer activations.
\item \textbf{Linear decodability} as \(R^2\) of linear regression from the first \(n\) PCs (\(n=1,\dots,30\)) to target digit \(t\).
\item \textbf{Generalization} as the accuracy of the pre-trained classifier over the held-out data image-action pairs.
\end{enumerate}

\begin{figure}[h]
\begin{center}
\includegraphics[width=1\textwidth]{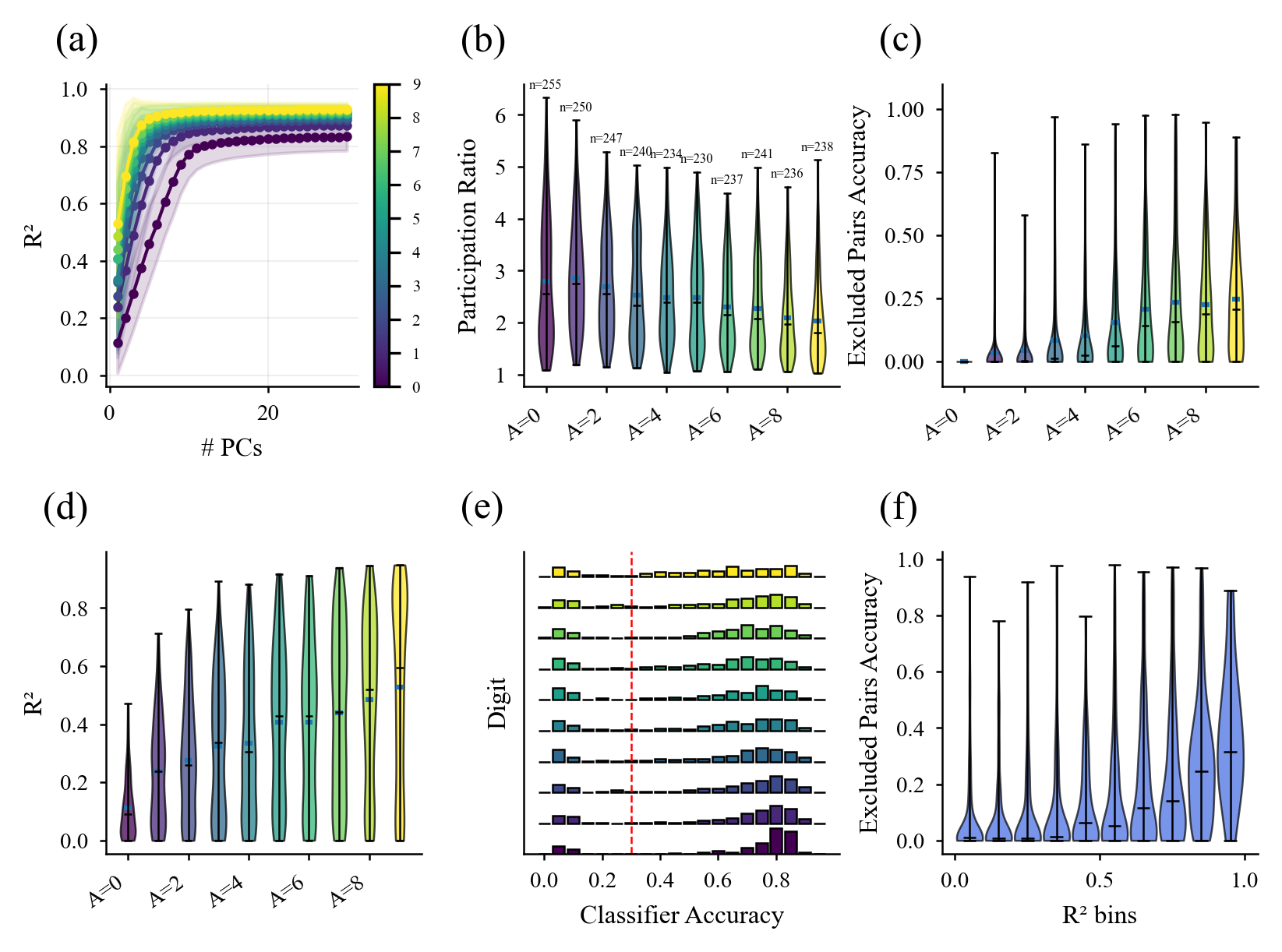}
\end{center}
\caption{MNIST predictive-learning analysis across prediction horizons.
Each point/violin entry corresponds to one independently trained model; horizon is indexed by maximal action \(A\). (a) Mean \(\pm\) s.d. (across runs) of target-label decodability (\(R^2\)) as a function of the number of retained principal components (PCs) of the learned representation, shown separately for each \(A\). (b) Distribution of representation PR across runs for each \(A\), quantifying effective dimensionality. (c) Distribution of excluded-pairs accuracy across runs for each \(A\). (d) Distribution of target-label decodability (\(R^2\); across runs for each \(A\)). (e) Per-digit classifier-accuracy distributions over generated samples (stacked histograms), showing class-dependent variability in output quality. Dashed red line shows cutoff used for inclusion in results. (f) Relationship between representation linearity and compositional generalization: excluded-pairs accuracy distributions (violins) after binning runs by maximal \(R^2\) (bin width \(0.1\)).
}
\label{fig_mnist_sup}
\end{figure}

Note that because GANs are notoriously variable and fragile, we trained many such models for each condition. We measured the pre-trained classifier accuracy over each trained model, and discarded models below a certain threshold as they are most likely models that converge to mode collapse. The threshold used is obtained from Figure~\ref{fig_mnist_sup}e.

\paragraph{Implementation and reproducibility notes.}
The implementation supports multi-process execution across GPUs.

\section{Sequential prediction}
\label{seq_appendix}
To test whether our results are not limited to settings where actions are encoded in action magnitude, we devise a minimal parallel abstract task based on sequential prediction \cite{levenstein2024sequential}.
This section documents the technical details and results.

\paragraph{Environment and sequence construction.}
We used a 1D corridor with \(S=30\) discrete states, and actions \(a \in \{-1,0,1\}\) encoded as one-hot vectors.
For each sequence length \(k \in \{1,\dots,30\}\), we generated all valid
state-action starts \((s,a)\) satisfying \(0 \le s+a < S\).
The input at the first step is
\[
x_1 = [\mathrm{onehot}(s),\mathrm{onehot}(a)].
\]
Subsequent inputs are action-only,
\[
x_t = [\mathbf{0},\mathrm{onehot}(a_t)], \quad t=2,\dots,k,
\]
where \(a_t\) is initially the same action as at \(t=1\).
If applying \(a_t\) would leave the corridor at later steps, the action is clamped to
zero for the remaining rollout (stay in place).
Targets are one-hot next states at each step, producing a sequence-labeling objective of length \(k\).

\paragraph{RNN architecture.}
The model has \(L=1\) recurrent layers, hidden size \(H=40\), and no biases.
Each layer uses an additive linear recurrence:
\[
h_t^{(\ell)} = W_{ih}^{(\ell)} z_t^{(\ell)} + h_{t-1}^{(\ell)},
\]
with \(z_t^{(1)}=x_t\) and \(z_t^{(\ell)}=h_t^{(\ell-1)}\) for \(\ell>1\).
No nonlinearity is applied between layers or across time.
Readout is a linear projection from the top hidden layer to \(S\) logits at each time step.

\paragraph{Optimization.}
Training uses Adam (\(\mathrm{lr}=10^{-3}\)) for 10,000 epochs on the full generated set for each \(k\)
(full-batch updates).
The loss is cross-entropy.
Accuracy is reported as the mean fraction of correct state predictions across all samples and time steps.

\begin{figure}[h]
\begin{center}
\includegraphics[width=1\textwidth]{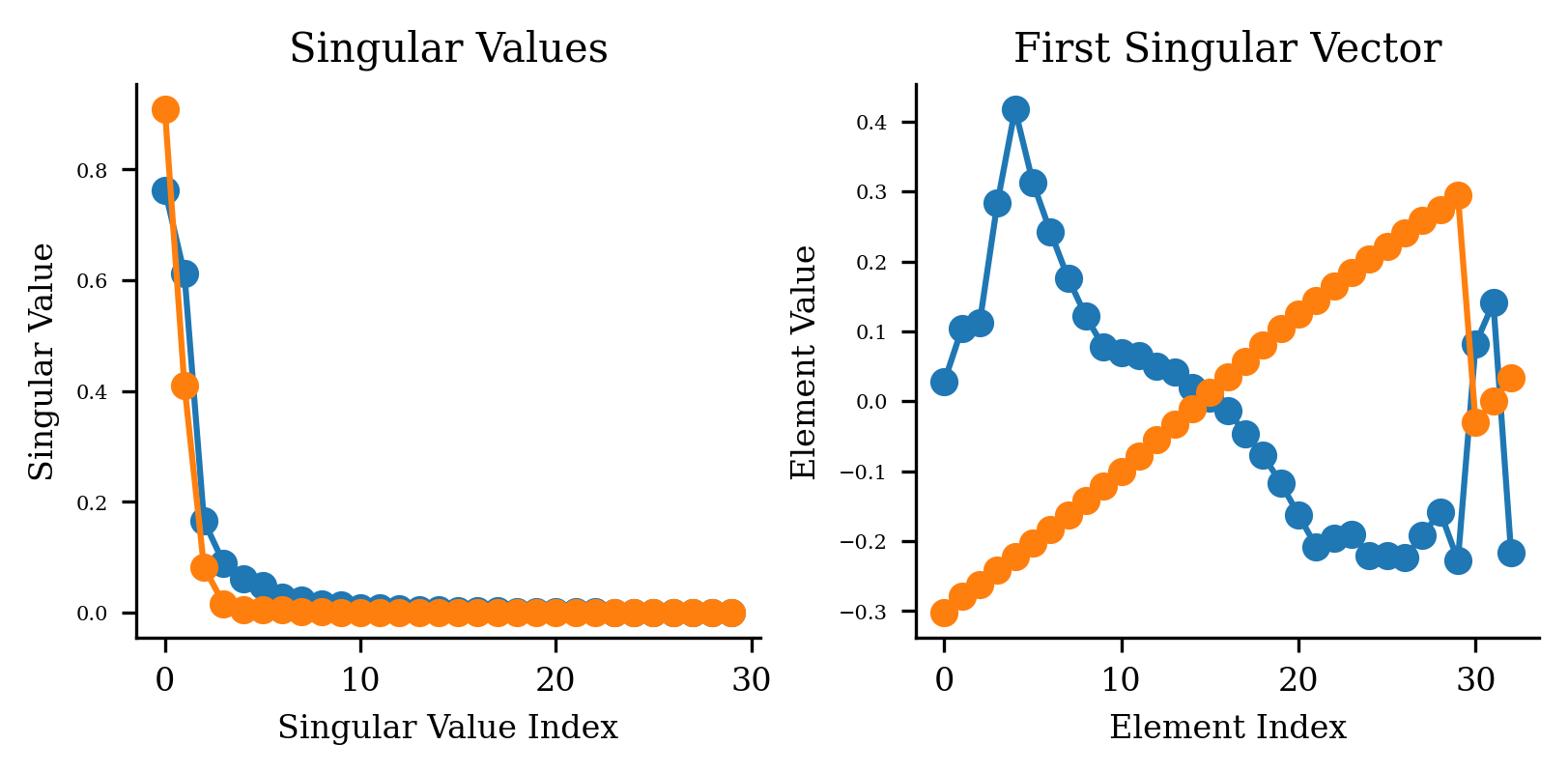}
\end{center}
\caption{Result of sequential actions experiments. We compare the singular value spectrum and first singular vector of the resulting weight matrix for two different prediction conditions. As can be seen, only when the prediction horizon is large ($k$, length of prediction sequence), the first singular vector corresponds to the linear transformation that extracts the scalar value of the output state.}
\label{fig_seq}
\end{figure}

\section{Predictive Coding Networks}
We summarize here the exact simulation protocol for the PCN experiments in Figure~\ref{fig_pcn}.

\paragraph{Task and dataset construction.}
For each run, we instantiate the abstract corridor environment with
30 states and one-hot state and action encodings.
The prediction horizon \(A\) was swept over
\(A \in \{1,\dots,29\}\).
For each \((s,a)\) pair, the model input is the concatenation of current-state and action encodings,
and the target is the next state encoding. 

\paragraph{Model architecture.}
We use a deep predictive coding network with \(L=5\) trainable layers and hidden widths
\([256,256,256,256]\), i.e., input \(\rightarrow\) 4 hidden layers \(\rightarrow\) output.
Weights are initialized as \(W_l \sim \mathcal{N}(0, \sigma_l^2)\) with
\(\sigma_l^2 = 2/(n_{\mathrm{in}}+n_{\mathrm{out}})\), and biases are initialized to zero.
Hidden layers use ReLU activations; the output layer uses a sigmoid nonlinearity.
Input features are standardized per dimension (zero mean, unit variance; near-zero standard
deviations are clamped to 1).

\paragraph{Predictive-coding inference and local learning.}
For each minibatch, latent states are first initialized by a feedforward pass, then relaxed for
\texttt{inference\_steps = 30} iterations using local prediction errors.
Denoting layer states by \(h_l\), prediction errors by
\(e_{l+1} = h_{l+1} - \hat{h}_{l+1}\), and activation function by \(\phi\),
hidden-state updates follow
\[
h_l \leftarrow h_l + \eta_{\mathrm{infer}}
\Big((e_{l+1}W_l)\odot \phi'(h_l) - e_l\Big),
\]
with \(\eta_{\mathrm{infer}} = 0.05\).
The output state is updated with a supervised mismatch term
\((h_L - y)\) in addition to the local prediction error.
After inference, synapses are updated with a local Hebbian-like rule
\[
\Delta W_l \propto e_{l+1}^{\top}\phi(h_l), \qquad
\Delta b_l \propto \mathrm{mean}(e_{l+1}),
\]
using \(\eta_{\mathrm{weight}} = 0.2\), without backpropagation through time or
autograd-based gradient propagation.

\paragraph{Optimization protocol.}
Each run uses full-batch training (\texttt{batch\_size = N}, where \(N\) is the full dataset size)
for 1000 epochs; 
We evaluate training accuracy and verify all simulations reached $100\%$ accuracy.
To quantify geometric organization, we compute (i) \(R^2\) between the first principal axis of
last-hidden activations and state labels, and (ii) Participation Ratio of the same activations.
For the PCA metric, evaluation is restricted to samples with \(|a|\le 1\).

\paragraph{Sweep and aggregation.}
We run 10 random seeds for each horizon value, producing \(29 \times 10 = 290\) simulations total.
Final plots report mean and standard deviation across seeds as a function of \(A\).

\section{Abstract task with binary tree geometry}

This section documents the exact data construction, model, and sweep protocol used
for the abstract non-Euclidean task in Figure~\ref{fig_tree}.

\paragraph{State space and action set.}
We define a rooted binary tree of depth \(d\), with states indexed as heap nodes
\(\{1,\dots,2^d-1\}\). The action set is
\(\mathcal{A}=\{\texttt{up},\texttt{down-left},\texttt{down-right}\}\),
implemented as integer actions \(\{0,1,2\}\), with one-hot encodings.
Transitions follow
\[
\texttt{up}: s \mapsto \lfloor s/2 \rfloor,\quad
\texttt{down-left}: s \mapsto 2s,\quad
\texttt{down-right}: s \mapsto 2s+1.
\]
Invalid transitions are excluded by construction.

\paragraph{Input-output construction for horizon \(k\).}
For each start-target pair \((s_{\mathrm{start}}, s_{\mathrm{target}})\), we compute the
shortest path on the tree via the lowest common ancestor decomposition:
move up from start to the ancestor, then down to the target.
Pairs requiring more than \(k\) moves are discarded.
The model input is
\[
x = [\mathrm{onehot}(s_{\mathrm{start}}), \mathrm{onehot}(a_1),\dots,\mathrm{onehot}(a_m), 0,\dots,0],
\]
where \(m\le k\) is the path length and the action segment is zero-padded to length \(k\).
The target is a one-hot code of the reached endpoint \(s_{\mathrm{target}}\).

\paragraph{OOD split protocol.}
To probe compositional generalization, we create held-out samples by omitting a subset of examples
that involve the maximal state index \(s_{\max}=2^d-1\).
Specifically, we collect all rows with \(s_{\mathrm{start}}=s_{\max}\) or
\(s_{\mathrm{target}}=s_{\max}\), then sample half of them uniformly at random into the test set.
The remaining rows form the training set.

\paragraph{Feedforward model.}
We train a ReLU multilayer perceptron (\texttt{DeepNet}) with:
\begin{itemize}
\item hidden width \(H\) (default \(H=512\)),
\item \(L\) hidden layers (swept; defaults include \(L\in\{1,2,3\}\)),
\item Xavier-normal initialization with gain \texttt{init\_scale},
\end{itemize}
The network outputs class logits over all tree states, and we optimize multiclass
cross-entropy against target state indices.

\paragraph{Optimization details.}
Training uses full-batch gradient descent with SGD for a fixed number of epochs
(default 100,000 per task). The optimizer is initialized with effective learning rate
\(\eta_0=\texttt{lr}/(2^L+1)\).
An adaptive schedule then updates the current learning rate between
\([\texttt{lr\_min},\texttt{lr\_max}]\): it is increased after sustained improvement and
decreased after sustained non-improvement, using configurable patience and multiplicative factors.

\paragraph{Sweep protocol and parallelization.}
For each sweep value and each seed, we run horizons
\[
k \in \{1,\dots,2(d-1)+1\}.
\]
With default \(d=6\), this gives \(k=1,\dots,11\).
Each task logs train/test loss and accuracy, final learning rate, and the full loss curve.

\newpage

\end{document}